\documentclass{article}

\PassOptionsToPackage{numbers, compress}{natbib}

\usepackage[eandd, final]{neurips_2026}

\usepackage[utf8]{inputenc} 
\usepackage[T1]{fontenc}    
\usepackage[hidelinks]{hyperref}       
\usepackage{url}            
\usepackage{booktabs}       
\usepackage{amsfonts}       
\usepackage{nicefrac}       
\usepackage{microtype}      
\usepackage{xcolor}         
\usepackage{graphicx}
\usepackage{booktabs}
\usepackage{amsmath}
\usepackage{amssymb}
\usepackage{microtype}
\usepackage{xcolor}
\usepackage{svg}
\usepackage{comment}
\usepackage{adjustbox}
\usepackage{tabularx}
\usepackage[inline]{enumitem}
\usepackage{cprotect}
\usepackage[noorphans,vskip=0.5ex]{quoting}
\usepackage{longtable}
\usepackage{makecell} 
\usepackage{pifont}
\usepackage{multirow}
\usepackage{wrapfig}
\usepackage{ragged2e}
\usepackage{siunitx}
\usepackage{subcaption}
\usepackage{tcolorbox}
\usepackage{array}
\usepackage{siunitx}
\usepackage{capt-of}
\usepackage{arydshln}
\usepackage{xspace}

\usepackage[symbol]{footmisc}

\usepackage[capitalize]{cleveref}
\crefname{section}{Sec.}{Secs.}
\Crefname{section}{Section}{Sections}
\Crefname{table}{Table}{Tables}
\crefname{table}{Tab.}{Tabs.}

\newcommand{\myparagraph}[1]{%
  \par\noindent\textbf{#1}\enspace\ignorespaces
}

\newcolumntype{Y}{>{\centering\arraybackslash}X}

\definecolor{turquoise}{cmyk}{0.65,0,0.1,0.3}
\definecolor{purple}{rgb}{0.65,0,0.65}

\newcolumntype{L}{>{\raggedright\arraybackslash}X}

\definecolor{fixmecolor}{rgb}{0.1,0.1,0.9} 
\definecolor{notecolor}{rgb}{0.1,0.7,0.1} 

\newcommand{\benchmark}{\textsc{Ego-MC-Bench}}
\newcommand{\augmentation}{\textsc{Ego-CoMist}}

\newcommand{\ie}{i.e.\@\xspace}
\newcommand{\eg}{e.g.\@\xspace}

\newcolumntype{C}[1]{>{\centering\arraybackslash}X[#1]}

\title{Streaming Interventions: Can Video Large Language Models Correct Mistakes as They Occur?}

\author{%
  Apratim Bhattacharyya\textsuperscript{1}\hspace{0.35cm} 
  Shweta Mahajan\textsuperscript{2, 3}\hspace{0.35cm} 
  Sanjay Haresh\textsuperscript{1}\hspace{0.35cm} \\
  \textbf{Rajeev Yasarla}\textsuperscript{1} \hspace{0.35cm}
  \textbf{Reza Pourreza}\textsuperscript{1}\hspace{0.35cm}
  \textbf{Litian Liu}\textsuperscript{1}\hspace{0.35cm}
  \textbf{Risheek Garrepalli}\textsuperscript{1} \hspace{0.35cm} \\
  \textbf{Roland Memisevic}\textsuperscript{1} \\
  \textsuperscript{1} Qualcomm AI Research\thanks{Qualcomm AI Research is an initiative of Qualcomm Technologies, Inc.}\hspace{0.35cm} \textsuperscript{2} York University\hspace{0.15cm} \textsuperscript{3} Vector Institute for AI
}

\begin{document}

\maketitle

\begin{abstract}
    Learning everyday skills, like cooking a dish, relies increasingly on instructional media such as online videos. 
    This opens the door to the use of video (and multimodal) large language models (LLMs) as task guidance assistants. 
    A crucial capability for the real-world success of a prospective task guidance assistant is its ability to intervene proactively as soon as a mistake is 
    apparent in order to guide the user.
    To evaluate this crucial capability, we introduce \benchmark{} (Mistake Corrections), a benchmark for evaluating \emph{reactive, step-by-step} task guidance in realistic cooking scenarios.
    Extensive experiments show that \benchmark{} is highly challenging for state‑of‑the‑art video LLMs.
    We argue that a key reason is the limited availability of training data for 
    fine-tuning models on this task. 
    Although there exists a wide range of cooking video datasets, existing datasets 
    lack examples of mistakes along with appropriately timed interventions. 
    To help address this data limitation, we also introduce \augmentation{}, a counterfactual synthetic dataset created by transforming non‑interactive cooking videos into supervised training examples showing proactive interventions.  
    We show that fine-tuning on \augmentation{} yields performance gains especially for smaller and more efficient video LLMs that are well suited for delivering assistance on edge devices.
\end{abstract}
%

\section{Introduction}
\label{sec:intro}
Traditionally, learning an everyday skill, such as preparing a new dish, requires 
reading a recipe or watching a video. 
Although being taught by domain experts, such as a chef, would be preferable, this
option is typically not available or too costly. 
Advances in multimodal large language models (LLMs) now allow AI systems to understand and respond to speech, audio, and visual inputs in real time~\citep{gpt52,abs-2507-06261,team2024gemini,xu2025qwen3}. This creates an opportunity to leverage such models for live, step‑by‑step guidance by emulating domain experts.

\begin{figure}[t!]
    \centering
    \includegraphics[width=0.97\linewidth]{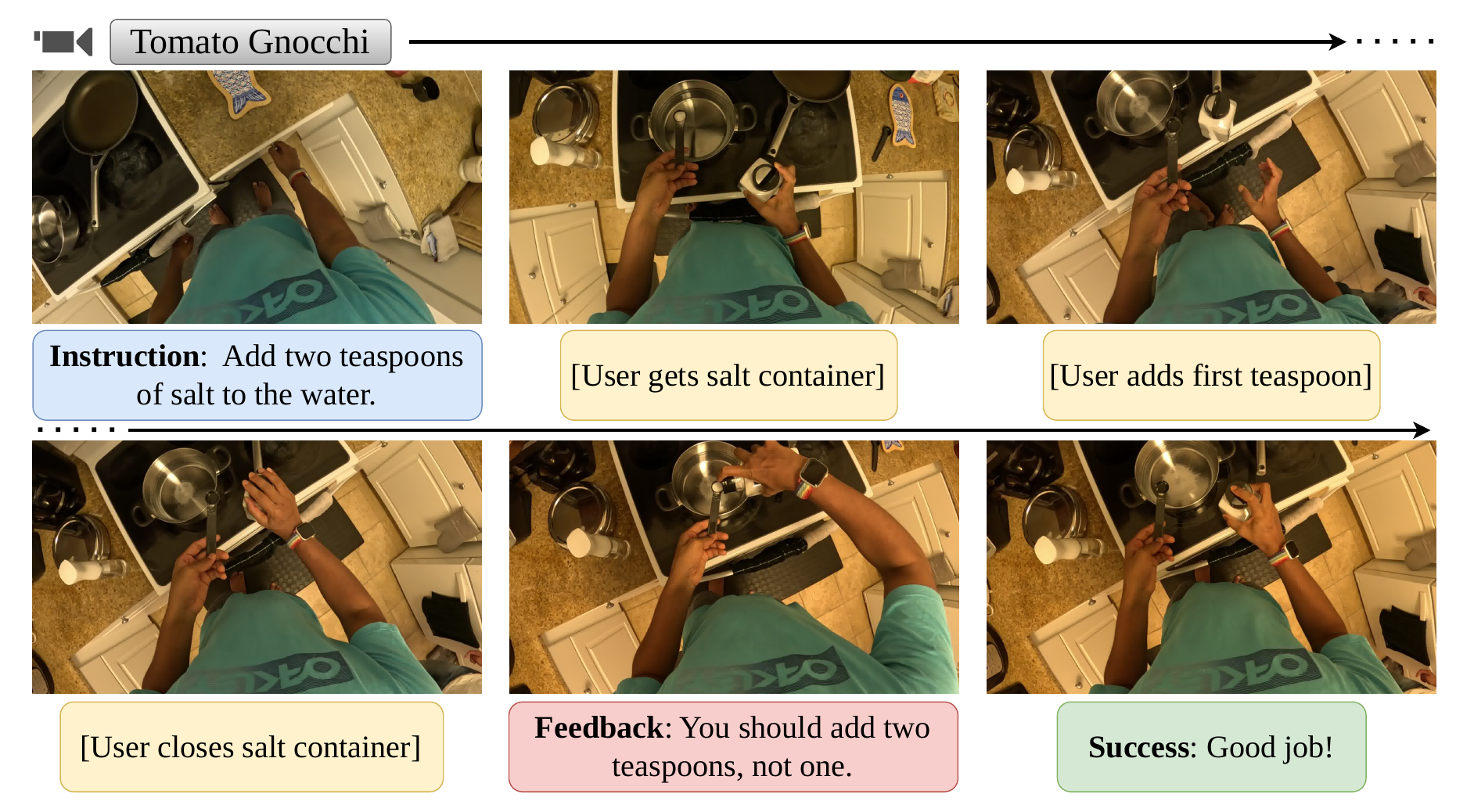} \\
    \vspace{-0.2cm}
    \includegraphics[width=0.97\linewidth]{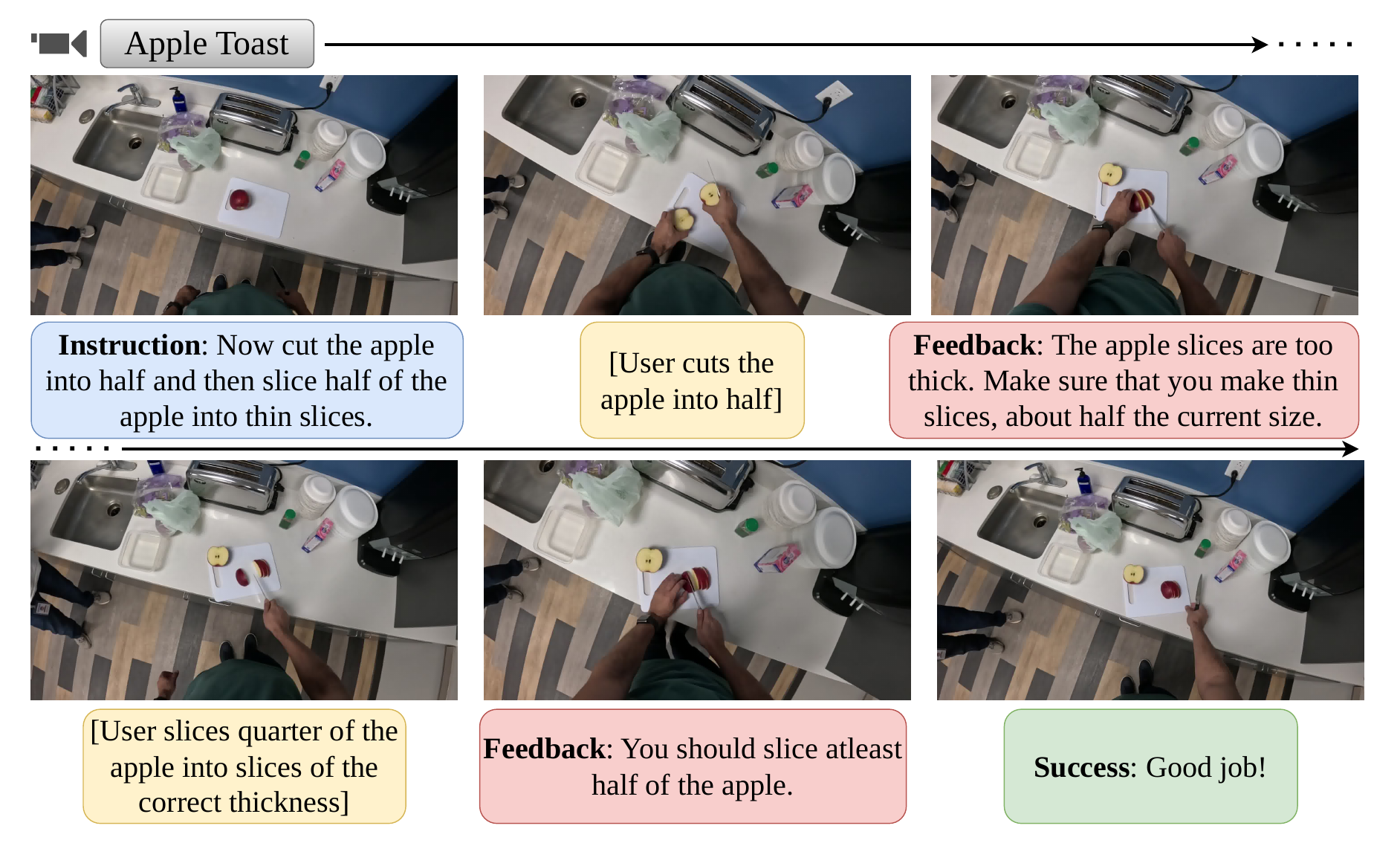}
    \vspace{-0.30cm}
    \caption{Our \benchmark{}: interventions with appropriate feedback whenever a mistake is apparent, guiding the user towards successful goal completion across recipe steps.}
    \label{fig:benchmark_examples}
\end{figure}

For multimodal and video LLMs to act as true multimodal assistants, they need to react \emph{proactively} to events in live video streams. Consider the example in \cref{fig:benchmark_examples} (top), where a person prepares Tomato Gnocchi. The recipe calls for two teaspoons of salt. 
The model should perceive that only one teaspoon has been added, 
infer from the video that the user does not intend to add more, 
and proactively prompt them to add the second teaspoon. Once the user does so, 
the model should detect and signal completion of this step.  
However, current benchmarks in this area~\citep{qicd_2025,WangKRPCABFTFJP23,wang2025topdownreasoningexplainablemultiagent} either test a limited subset of conversational abilities or re‑purpose non‑reactive data—providing only a partial assessment of proactive assistance capabilities of multimodal and video LLMs.

Existing attempts to build assistants for proactive task guidance in the real world have been met with limited success~\cite{qicd_2025,ZhangDLMKDCM25}. 
This is in spite of the fact that there are many large scale datasets in the cooking domain~\cite{Damen2022RESCALING,grauman2023ego,PeddiACPVGZWKRR24,zhou2018towards}. 
The key challenge is that such datasets lack suitable demonstrations of mistakes and corresponding interventions and feedbacks. Furthermore, collecting such high-quality supervision at scale is prohibitively expensive.

In this work, we introduce a novel benchmark and data generation pipeline to help address these issues. 
Our contributions in detail are,
\begin{enumerate*}
    \item We propose the Qualcomm Interactive Cooking: \benchmark{} (Mistake Corrections)\footnote{Project page -- available at this \href{https://apratimbh.github.io/livecookv2/}{URL}.}, the first benchmark to evaluate the step-by-step task guidance capabilities of multimodal LLMs in a truly reactive setup. 
    \benchmark{} evaluates the ability of a model to intervene as soon as a mistake is apparent by providing appropriate feedback and thus guiding users to successful task completion.
    \item We propose Qualcomm Interactive Cooking: \augmentation{}, a novel synthetic dataset created by re-purposing existing (non-interactive) datasets towards supervised training on proactive, streaming task guidance.
    Overall, this represents a significant expansion of the Qualcomm Interactive Cooking Dataset~\cite{qicd_2025}.
    \item We present extensive experiments which demonstrate that the interactive task guidance task in our \benchmark{} is highly challenging for state of the art video LLMs.
    \item We show that fine-tuning a model on the \augmentation{} dataset leads to significant improvements in successful mistake interventions, especially for efficient video LLMs well suited for edge-deployment.
\end{enumerate*}

\begin{table}[t!]
    \centering
    \footnotesize
    \setlength{\tabcolsep}{3pt}
    \renewcommand{\arraystretch}{1.08}
    \caption{Here, \emph{Multi-step Goal Driven} refers to whether the videos are driven by a specific goal (\eg, cooking a recipe); \emph{Step-by-Step Instructions}: whether the videos contain participants following a set of step-by-step instructions; \emph{Timed Feedback}: whether the participants receive timed feedback per step that is sufficient to guide the subjects to the goal; \emph{Reactive Participants}: whether the participants react to the feedback and correct their actions to successfully complete the complex multi-step goal.}
    \label{tab:dataset_comp}

    \begin{tabular*}{\linewidth}{@{\extracolsep{\fill}} l l c c c c c @{}}
        \toprule
        Dataset & Domain &
        \makecell[c]{Multi-step\\Goal Driven} &
        \makecell[c]{Step-by-Step\\Instructions} &
        \makecell[c]{Timed\\Feedback} &
        \makecell[c]{Reactive\\Participants} &
        \makecell[c]{Length\\(hrs)} \\
        \midrule
        Assembly-101~\cite{SenerCSHSWY22}      & Toy Assembly & \checkmark & $\times$   & $\times$   & $\times$ & 513 \\
        HowTo100M~\cite{miech2019howto100m}    & Diverse      & \checkmark & \checkmark & $\times$   & $\times$ & 134k \\
        COIN~\cite{TangDRZZZL019}              & Diverse      & \checkmark & \checkmark & $\times$   & $\times$ & 512 \\
        YouCookv2~\cite{zhou2018towards}       & Cooking      & \checkmark & \checkmark & $\times$   & $\times$ & 176 \\
        WTAG~\cite{BaoYZSBISZC23}              & Cooking      & \checkmark & \checkmark & $\times$   & $\times$ & 10 \\
        HoloAssist~\cite{WangKRPCABFTFJP23}    & Obj. manip.  & $\times$ & \checkmark   & \checkmark   & \checkmark & 166 \\
        QEVD~\cite{PanchalBB0BD0LM24}          & Fitness      & $\times$   & \checkmark & \checkmark & \checkmark & 474 \\
        QICD~\cite{qicd_2025}                                  & Cooking      & \checkmark & \checkmark & \checkmark & $\times$ & 94 \\
        \midrule
        \augmentation{} (Ours)                 & Cooking      & \checkmark & \checkmark & \checkmark & $\times$ & 124 \\
        \benchmark{} (Ours)                    & Cooking      & \checkmark & \checkmark & \checkmark & \checkmark & 10 \\
        \bottomrule
    \end{tabular*}
\end{table}

\section{Related Work}
\myparagraph{Multimodal and Video LLMs.} Vision–language models have advanced rapidly on the heels of breakthroughs in large‑scale language modeling. Early work such as Flamingo~\cite{alayrac2022flamingo} paired a pretrained language model with learnable vision adapters, aligning visual and linguistic representations to enable strong zero‑ and few‑shot performance. This paradigm has since yielded a diverse family of large multimodal language models~\cite{achiam2023gpt, bai2023qwen, abs-2511-21631, grattafiori2024llama, team2023gemini, team2024gemini, Qwen2VL} with strong results on video understanding. Progress has been further catalyzed by broad evaluation suites such as Video‑MME~\cite{fu2025video}, Video‑MMMU~\cite{abs-2501-13826}, and LongVideoBench~\cite{WuLCL24}. 
However, most existing models and benchmarks remain largely \emph{turn‑based}: 
responses only when explicitly prompted instead of proactively monitoring,  
or reacting to, the unfolding events in the stream. 
In contrast, our \benchmark{} benchmark and \augmentation{} synthetic data generation pipelines are designed for \emph{interactive} scenarios, where a model must track events in its input video stream and respond without explicit user prompts.

\myparagraph{Streaming Video LLMs.} Standard video LLMs struggle with the temporal demands of streaming inputs, motivating recent models that respond interactively to live video. A foundational effort is {VideoLLM‑online}~\cite{VideoLLM-online}, trained to narrate egocentric streams. Building on this, works such as {StreamMind}~\cite{ding2025streammind}, {Flash‑VStream}~\cite{zhang2024flash}, {LiveVLM}~\cite{abs-2505-15269}, and {ReKV}~\cite{di2025rekv} propose more efficient architectures for the streaming narration setting. {StreamingVLM}~\cite{abs-2510-09608} extends narration to hour‑long videos by using attention sinks, while {LION‑FS}~\cite{LION-FS} adopts a two‑stream fast–slow vision encoder to improve visual grounding and accuracy. 
A different interaction scenario is considered by {ViSpeak}~\cite{fu2025vispeak}, 
which enables models to respond to gestures and body language in real time. In contrast to narration or QA, we study the \emph{streaming intervention} task: detecting both mistakes and step completions within a procedural activity and intervene at the right moment, which requires fine‑grained tracking of progress toward the target goal.

\myparagraph{Benchmarks and Datasets for Streaming Video.}
VideoLLM-online~\cite{VideoLLM-online} introduced an Ego4D‑based narration dataset instrumental in the development of streaming video LLMs. LiveCC~\cite{chen2025livecc} contributed ASR‑aligned datasets for live sports commentary. ProactiveQA~\cite{ZhangDLMKDCM25} provides a streaming dialogue and question‑answering dataset over egocentric videos. SVBench~\cite{yang2025svbench}, OmniNMI~\cite{omnimmi}, Ovo‑Bench~\cite{li2025ovo}, and Streamo~\cite{xia2025streaming} target streaming dialogue and QA over general, real‑world videos. By contrast, datasets such as HoloAssist~\cite{WangKRPCABFTFJP23}, WTAG~\cite{BaoYZSBISZC23}, and QEVD~\cite{panchal2024saysayitlive} are designed for interactive assistance that delivers live feedback in daily‑life, cooking, and fitness scenarios. However, unlike our \benchmark{} benchmark, these datasets do not provide multi‑step feedback that guides users toward successful completion of complex goals as shown in \cref{tab:dataset_comp}. Finally, the base Qualcomm Interactive Cooking Dataset~\cite{qicd_2025} (QICD) frames step‑by‑step task guidance in a non‑reactive setting. In contrast, \benchmark{} is explicitly \emph{reactive}, evaluating interventions to mistakes as they occur.


\begin{figure}[t]
  \centering
  \begin{minipage}[t]{0.48\textwidth}
    \vspace{0pt}
    \centering
    \includegraphics[width=\linewidth]{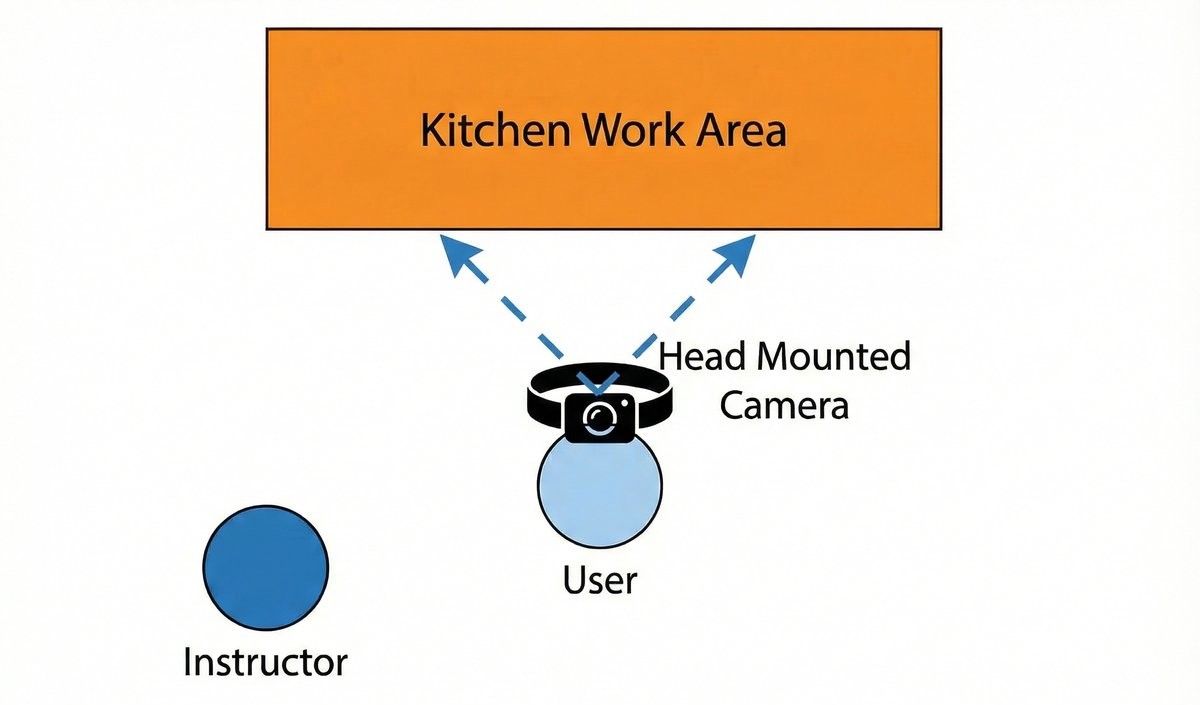}
    \caption{Recording setup: Dashed lines show the camera's field of view.}
    \label{fig:recording_setup}
  \end{minipage}
  \hfill
  \begin{minipage}[t]{0.48\textwidth}
    \vspace{0pt}
    \centering
    \includegraphics[width=\linewidth]{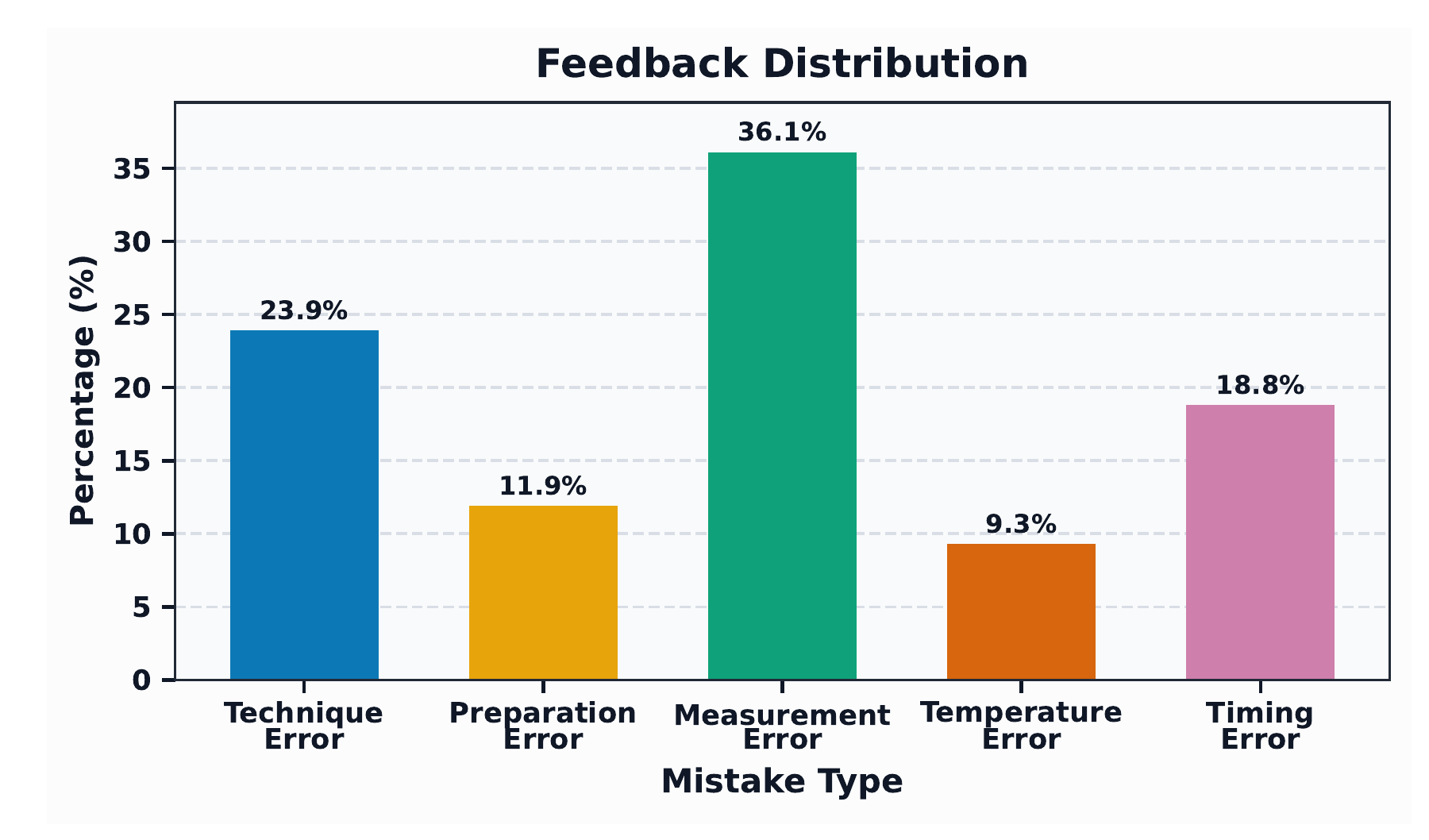}
    \caption{Distribution of feedbacks in \benchmark{} using classification of~\cite{PeddiACPVGZWKRR24}.}
    \label{fig:mistake_dist}
  \end{minipage}
\end{figure}

\section{\benchmark{}}
We introduce Qualcomm Interactive Cooking: \benchmark{} (\cref{fig:benchmark_examples}), a benchmark for evaluating multi‑modal assistants that guide users \emph{step by step} through cooking tasks with multi‑step recipes. 
We focus on cooking because it offers a practical setting where timely intervention matters: mistakes such as adding too much salt can be difficult to reverse, making delayed feedback costly. Simple-to-moderate recipes require relatively little specialized knowledge, helping isolate the challenges of detecting mistakes and intervening at the right moment in streaming video.

\benchmark{} targets two core capabilities: (i) detecting step completion and (ii) intervening on mistakes with timely, corrective feedback (we define mistakes as \emph{actions that directly impede recipe completion}). Prior step‑by‑step benchmarks~\cite{qicd_2025} are derived from non‑interactive datasets (e.g., CaptainCook4D~\cite{PeddiACPVGZWKRR24}), producing non‑reactive scenarios where users cannot respond to feedback. In contrast, \benchmark{} models a realistic interactive setting: users react to feedback, and the assistant actively guides them toward successful completion. Next, we prodvide a a more formal definition of mistakes in context of our \benchmark{} benchmark.

\myparagraph{Mistakes.} We define mistakes for procedural tasks such as cooking to be any action that hinders the successful completion of the given instruction (goal) within a reasonable amount of time. Measurement, timing, temperature errors are quite obvious examples that would lead to either a poor quality outcome or a completely inedible dish.

\myparagraph{Benchmark Collection.} The \benchmark{} benchmark is recorded in an interactive live setup. The recording is performed using a head mounted 
camera in a kitchen. An instructor provides step by step instructions and feedback. The step by step instructions are recipe steps of varying  complexity (\cref{fig:benchmark_examples}).
 The instructor is positioned behind the participant as shown in \cref{fig:recording_setup} such that the instructor can observe the user actions in detail while not being in the field of view of the head mounted camera. The participants are not provided with the recipe beforehand. This setup simulates real-world scenarios where an assistant guides a user step by step through a recipe. 
 To create a challenging benchmark, we intentionally include a broad range of realistic user errors. 
 Prior to each recording session, participants are coached on common mistake types using practice recipes that differ from the one being filmed. We then run brief mock recording sessions, in which participants deliberately make mistakes and receive targeted feedback on the plausibility and naturalness of those mistakes. Finally, after each session, every video is manually inspected to ensure quality and realism.

\myparagraph{Annotation and Verification.} \benchmark{} features manually generated variants of everyday recipes, deliberately limiting the influence of prior knowledge so that evaluation emphasizes visual reasoning. The annotations in our \benchmark{} benchmark consist of instructions and feedbacks, provided by the instructor to the participant as described above. The voice recordings are transcribed and then manually verified by an independent annotator to ensure accuracy. Additionally, the annotators are tasked with classifying the transcripts into instructions or feedbacks.  

\myparagraph{Realism and mistake diversity.} The benchmark collection protocol described above is designed for both realism and mistake diversity. The important steps in the protocol which ensures realism and mistake diversity are,
\begin{enumerate*}
\item \emph{Unseen recipes}: The participants do not know the recipe being recorded beforehand.
\item \emph{Open-ended participant coaching}: Coaching introduces general mistake categories using different recipes, without specifying which mistakes to make or when.
\item \emph{Realism-focused mock sessions}: In the mock recording sessions, the participants are provided with feedback to ensure that mistakes are plausible and natural, rather than rehearsing specific actions.
\item \emph{Final manual inspection}: Finally, if any mistake appears to be overly scripted or unrealistic then the video is discarded. 
\end{enumerate*}

\myparagraph{Benchmark statistics.} The benchmark contains $\sim$10 hours of video data across 40 recording sessions. It features 7 participants in total and includes diverse kitchen setups (\cref{fig:benchmark_examples}). It covers 559 recipe steps (across the 39 recording sessions) and 395 feedback messages corresponding to mistakes made by the participants (954 feedbacks in total). This also includes 22 clarification questions asked by the participants and subsequent clarifications provided by the instructor, along with 30 user comments usually acknowledging instructions or indicating preferences.

\benchmark{} includes two timing-based feedback types: \emph{anticipatory} feedback, given when a mistake is imminent (e.g., warning a user not to add too much salt before they pour), and \emph{post-error} feedback, given after a mistake has occurred when it could not have been anticipated (e.g., noticing that onions have already started to burn and advising the user to lower the heat). Irrecoverable mistakes appear only in the anticipatory setting, since once they occur, successful task completion is impossible.
\benchmark{} contains an approximately equal number of mistakes 
for each type (49.2\% vs 50.8\%).
This is very different from QICD~\cite{qicd_2025} where all feedbacks are \emph{post-error}.
To further highlight the diversity of mistakes in the \benchmark{} benchmark, we additionally include an analysis of the types of mistakes, using the classification scheme from CaptainCook4D~\cite{PeddiACPVGZWKRR24} in \cref{fig:mistake_dist}.

\myparagraph{Qualitative examples.} We show qualitative examples from \benchmark{} in \cref{fig:benchmark_examples}. In the top row, the assistant instructs the participant to add two teaspoons of salt; when only one is added, it intervenes with corrective feedback, after which the participant completes the step and receives confirmation. In the bottom row, the assistant instructs the participant to halve an apple and slice half of it thinly; it first corrects slice thickness and then the sliced quantity before the participant successfully completes the task. These examples illustrate the central challenge addressed by \benchmark{}: providing timely, appropriate feedback as soon as a mistake becomes apparent.

\section{\augmentation{}: Synthetic Counterfactual Mistakes}
As \benchmark{} evaluates models in a realistic fully interactive setting, it serves as a gold-standard benchmark for studying training methods and datasets. It's stringent downstream evaluation also enables more exploratory data generation, including (noisy) synthetic data, since any drawback or benefit is directly reflected in task performance. Here, we introduce Qualcomm Interactive Cooking: \augmentation{}, a counterfactual dataset of interactive instruction--feedback pairs. Built from existing non-interactive datasets~\cite{Grauman_2022_CVPR,grauman2023ego,PeddiACPVGZWKRR24}, it approximates the \benchmark{} setting by providing feedback at the earliest point a mistake becomes apparent. Given a video clip of a recipe step, \augmentation{} consists of two stages: (i) generating a counterfactual step with corresponding instruction and feedback, and (ii) inferring the timestamp at which the feedback should be delivered. We illustrate both stages in \cref{fig:counterfactual_generation_process} for measurement errors and provide details in the following.

\subsection{Stage 1: Counterfactual Instruction and Feedback Generation}
\label{sec:stage_1_ego_comist}
\begin{figure}[!t]
  \centering
  \includegraphics[width=\linewidth]{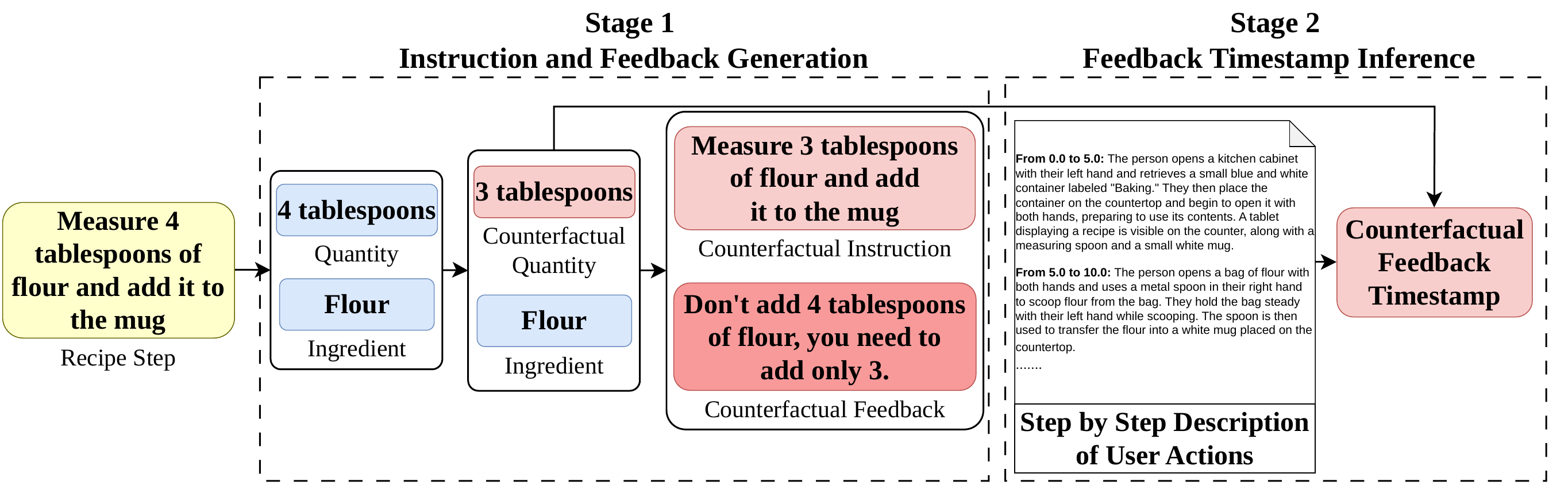}
  \vspace{-0.5cm}
  \captionof{figure}{Our \augmentation{} Counterfactual mistake generation pipeline for measurement errors.}
  \label{fig:counterfactual_generation_process}
\end{figure}

In the first stage, we synthesize counterfactual variants of each recipe step by perturbing specific attributes: ingredient quantity (measurement errors), preparation method (preparation errors), cooking technique (technique errors), duration (timing errors), and temperature/heat setting (temperature errors). Together, these categories capture the vast majority of common cooking mistakes \cite{PeddiACPVGZWKRR24}.
To this end, we extract the relevant attributes from the recipe step, \ie, quantity, cooking technique, preparation method, cooking time and cooking temperature, using a state of the art LLM (Gemini-2.5-Pro~\cite{team2024gemini}) as shown in \cref{fig:counterfactual_generation_process}.
We then generate the corresponding counterfactual attribute, \eg, a counterfactual measurement, preparation method, cooking technique, heat setting or cooking time.
Finally, given the original recipe step and the counterfactual attribute, we generate a counterfactual instruction and feedback pair. Next we describe this process in detail for the measurement error type and provide details of the remaining error types and prompts used in the appendix.

\begin{figure}[t]
  \centering
  \includegraphics[width=0.9\linewidth]{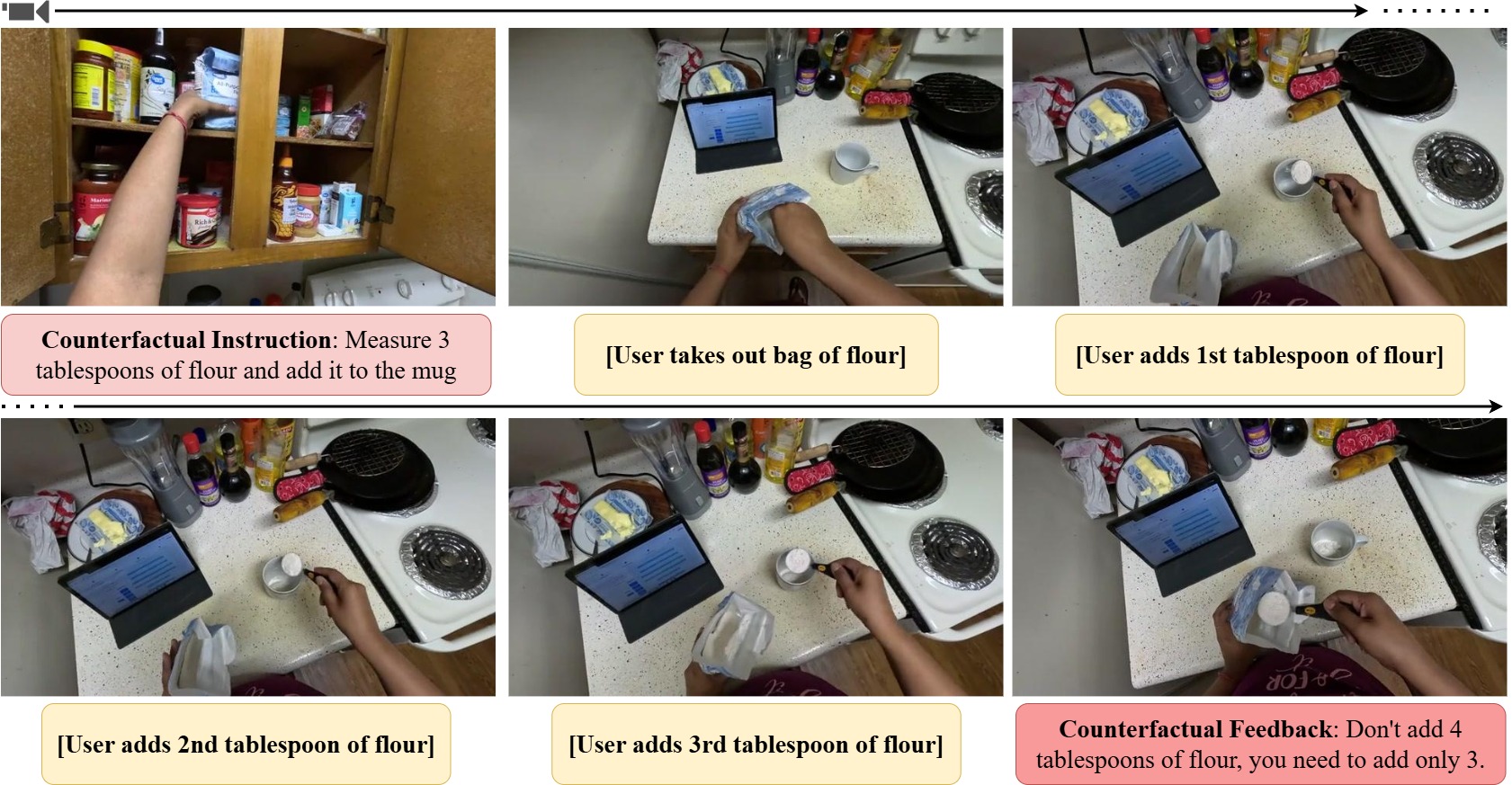}
  \vspace{-0.15cm}
  \captionof{figure}{Counterfactual mistake annotation in \augmentation{}.}
  \label{fig:counterfactual_ex1}
\end{figure}

\myparagraph{Measurement error.} Given a recipe step that requires a specific quantity of an ingredient(s), \eg, \emph{Measure 4 tablespoons of flour and add it to the mug}, we first ask the model to extract the specific quantity (\ie, \emph{4 tablespoons}) and the corresponding ingredient (\ie, \emph{flour}). Next, we ask the model to create a counterfactual quantity, that is reasonable for the given recipe step as shown in \cref{fig:counterfactual_generation_process}. In \cref{fig:counterfactual_ex1}, the model chooses 3 tablespoons as the counterfactual quantity. Then, given the original quantity and the counterfactual quantity, we ask the model to generate a new counterfactual instruction, \ie, \emph{Measure 3 tablespoons of flour and add it to the mug}. This allows us now to construct the feedback corresponding to this counterfactual instruction: \emph{Don't add 4 tablespoons of flour, you need to add only 3}. Additionally, to keep the dataset balanced, we generate an equal number of counterfactual instructions with smaller and larger measurements.


\subsection{Stage 2: Counterfactual Feedback Timestamp Inference}
\label{sec:stage_2_ego_comist}
At the second stage, we identify the appropriate timestamp for intervention and generate the corresponding (counterfactual) feedback. We first obtain step-by-step narrations of the input video clip. 
Given the original recipe step, the counterfactual recipe step, and the narration, we use a state-of-the-art LLM (Gemini-2.5-Pro) to infer when the feedback should be provided.
This strategy is effective because the feedback timestamp is generated in an \emph{oracle} setup, 
where the LLM has access to the counterfactual mistake and the step by step 
descriptions of the entire video clip, both of which are not available at inference time.

\myparagraph{Step-by-step descriptions.} To localize the intervention time, we require detailed, temporally aligned narrations focused on the user’s actions. When such descriptions are not already available in the dataset we generate them using a state-of-the-art video LLM (Qwen3-VL-32B-Instruct~\cite{abs-2511-21631}) using a sliding-window approach that prompts the video LLM to describe overlapping video chunks (10 sec windows with 5 sec stride). We instruct the model to focus on hand–object interactions, which are critical for identifying errors, \eg, detecting when the user begins adding an extra tablespoon of flour. Given these narrations, we infer the appropriate feedback timestamp using a state-of-the-art LLM. We next describe this process for the measurement error type and provide details of the remaining error types in the appendix.

\myparagraph{Measurement error.} When the counterfactual measurement is smaller than the original quantity in the recipe step, we prompt the model to find the timestamped description at which it becomes clear that the person intends to use more than the counterfactual amount. For example, in \cref{fig:counterfactual_ex1}, we ask the model to identify when the person is about to add a fourth tablespoon of flour, and use that timestamp for counterfactual feedback. When the counterfactual quantity is larger than the original quantity, we instead wait until it becomes clear that the person does not intend to add any more of the ingredient.

\subsection{\augmentation{}: Statistics and Human Evaluation}
\label{sec:aug_dataset}
Using the pipeline described above we 
generate counterfactual mistakes for the following existing cooking 
datasets: CaptainCook4D~\cite{PeddiACPVGZWKRR24}, Ego4D~\cite{GraumanWBCFGH0L22} and Ego-Exo4D~\cite{grauman2023ego}. 
As CaptainCook4D already contains mistakes 
we use the recipe steps in CaptainCook4D without mistakes in our \augmentation{} pipeline. 
In case of Ego4D and Ego-Exo4D, we leverage the Goal-Step~\cite{SongBNWMT23} and key step annotations, respectively, to filter for appropriate videos containing cooking activities. 
We use these annotations to obtain video descriptions and leverage them to generate instructions. However, these descriptions are less detailed compared to CaptainCook4D and do not usually mention details like cooking temperature, measurement or duration.  
We therefore only generate preparation and technique errors from these datasets.
Overall, \augmentation{} contains 4969, 13847, 6271 instruction-feedback pairs from CaptainCook4D, Ego4D and Ego-Exo4D respectively.

\begin{wrapfigure}[14]{r}{0.45\textwidth}
\vspace{-0.75cm}
  \begin{center}  \includegraphics[width=0.45\textwidth]{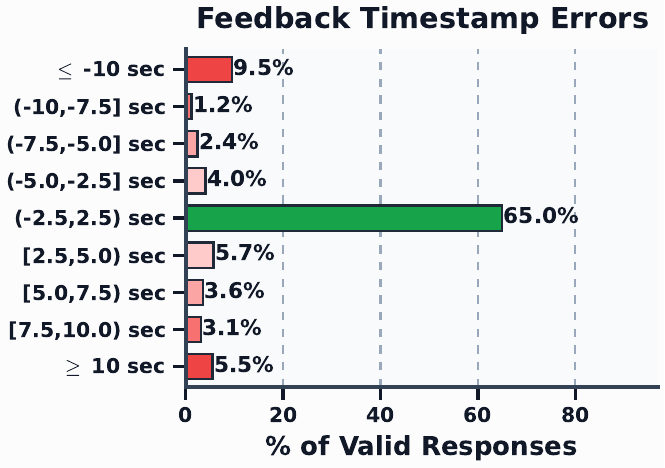}
  \end{center}
  \vspace{-0.35cm}
  \caption{\augmentation{}: Analysis of errors in timestamp inference (stage 2).}
  \label{fig:mistake_timestamp_dist}
\end{wrapfigure}
Finally, we perform a user study of the quality of the annotations in \augmentation{}. 
We randomly select 500 samples and first ask users to check if the example is valid. 
An example is invalid if the counterfactual instruction:
\begin{enumerate*}
    \item is not semantically different from the action in the clip, or,
    \item is not feasible with the current ingredients, or, 
    \item does not belong to the annotated mistake type.
\end{enumerate*}
The users found 87.1\% of instruction-feedback pairs to be valid.
Next, we asked them to find the appropriate time to provide the feedback, using the same criterion as for \benchmark{}.
\cref{fig:mistake_timestamp_dist} shows the resulting distribution of errors between 
the annotated feedback timestamp and the timestamp provided by the participants.
We see that most predictions ($\sim$ 65\%) are near-correct with error within $(-2.5,2.5)$ sec. Extreme deviations ($\geq$ 10 sec) remain a minority. Both ``on time'' annotations and annotations close to being on time (< 10 sec) provide useful training signals.

\begin{table}[t!]
    \centering
    \small
    \caption{Evaluation on the \benchmark{}, using the regular interval prompting strategy of \cite{qicd_2025}.} 
    \label{tab:streamcook_zeroshot} 

    \begin{tabularx}{\linewidth}{@{}
        >{\raggedright\arraybackslash}p{3.5cm} 
        >{\centering\arraybackslash}p{1.75cm}   
        @{\hspace{0.25cm}}
        >{\centering\arraybackslash}p{1.1cm}   
        >{\centering\arraybackslash}p{1.1cm}   
        >{\centering\arraybackslash}p{1.1cm}   
        >{\centering\arraybackslash\itshape}p{1.1cm}   
        >{\centering\arraybackslash\itshape}p{1.8cm}   
        @{}}
        \toprule
         & Instruction & \multicolumn{5}{c}{Mistake} \\
         \cmidrule{2-2} 
         \cmidrule{3-7}
        Method & {IC-Acc$\uparrow$} & Prec.$\uparrow$ & Rec.$\uparrow$ & F1$\uparrow$ & BERT$\uparrow$ & ROUGE-L$\uparrow$ \\
        \midrule
        \multicolumn{7}{c}{Per-recipe step} \\
        \midrule
        InternVL3.5-38B~\cite{wang2025internvl3_5} & 3.9 & 0.00 & 0.00 & 0.00 & 0.000 & 0.000\\
        Qwen2.5-VL-32B~\cite{abs-2502-13923} & 27.3 & 0.00 & 0.00 & 0.00 & 0.000 & 0.000 \\
        Qwen3-VL-8B~\cite{abs-2511-21631}  & 30.7 & 0.00 & 0.00 & 0.00 & 0.000 & 0.000\\
        VideoLLaMA3-7B~\cite{damonlpsg2025videollama3} &  31.8 & 0.00 & 0.00 & 0.00 & 0.000 & 0.000\\
        Qwen3.5-2B~\cite{qwen3p5} & 0.0 & 0.02 & 0.44 & 0.05 & 0.184 & 0.130\\
        Qwen3.5-9B~\cite{qwen3p5} & 6.1 & 0.07 & 0.53 & 0.12 & 0.201 & 0.137\\
        Qwen3.5-27B~\cite{qwen3p5} & {\bfseries 45.5} & {0.12} & {0.27} & {0.17} & 0.206 & 0.137\\
        Qwen3-VL-32B~\cite{abs-2511-21631} & 6.8 & 0.11 & {0.55} & {0.18} & 0.068 & 0.092 \\
        \midrule
        Videollm-online~\cite{VideoLLM-online} & 2.7 & 0.02 & 0.38 & 0.05 & 0.265 & 0.201\\
        LiveCC~\cite{chen2025livecc} & 1.6 & 0.03 & {\bfseries 0.43} & 0.06 & 0.248 & 0.196\\
        \midrule
        Gemini-2.5-Flash~\cite{abs-2507-06261} & 24.6 & 0.17 & 0.32 & {\bfseries 0.23} & 0.180 & 0.135 \\
        Gemini-3-Flash~\cite{gemini-flash}  & 32.7 & {\bfseries 0.19} & 0.30 & {\bfseries 0.23} & 0.126 & 0.102 \\
        \midrule
        \multicolumn{7}{c}{Full recipes} \\
        \midrule
        Qwen3-VL-32B~\cite{abs-2511-21631} & 6.8 & { 0.04} & {\bfseries 0.40} & { 0.07} & 0.061 & 0.090\\
        Qwen3.5-27B~\cite{qwen3p5} & {\bfseries 30.3} & 0.05 & 0.20 & 0.08 & 0.201 & 0.136\\
        \midrule
        Gemini-3-Flash~\cite{gemini-flash}  & 10.6 & {\bfseries 0.05} & {0.33} & {\bfseries 0.09} & 0.097 & 0.091\\
        \bottomrule
    \end{tabularx}
\end{table}

\begin{table}[t!]
    \centering
    \small
    \caption{Evaluation on our \benchmark{}. For the Qwen3-VL/3.5 models, we specify the dataset used for fine-tuning in brackets.} 
    \label{tab:cc_finetune_streamcook} 
    \small
    \begin{tabularx}{\linewidth}{@{}
        >{\raggedright\arraybackslash}p{4.2cm} 
        >{\centering\arraybackslash}p{1.5cm}   
        @{\hspace{0.2cm}}
        >{\centering\arraybackslash}p{1.1cm}   
        >{\centering\arraybackslash}p{1.1cm}   
        >{\centering\arraybackslash}p{1.1cm}   
        >{\centering\arraybackslash\itshape}p{1.05cm}   
        >{\centering\arraybackslash\itshape}p{1.6cm}   
        @{}}
        \toprule
         & Instruction & \multicolumn{5}{c}{Mistake} \\
         \cmidrule{2-2} 
         \cmidrule{3-7}
        Method & {IC-Acc$\uparrow$} & Prec.$\uparrow$ & Rec.$\uparrow$ & F1$\uparrow$ & BERT$\uparrow$ & ROUGE-L$\uparrow$ \\
        \toprule
       
        \multicolumn{7}{c}{{Per-recipe step}} \\
        \midrule
        ProAssist~\cite{ZhangDLMKDCM25} & 3.0 & 0.31 & 0.14 & 0.20 & 0.281 & 0.173\\
        {Qwen3.5-2B (QICD)} & 28.9 & {\bfseries 0.75} & 0.06 & 0.11 & 0.359 & 0.218 \\
        {Qwen3.5-2B (\augmentation{})} & 32.1 & 0.34 & 0.13 & 0.19 & 0.359 & 0.229\\
        {Qwen3-VL-2B (\augmentation{}+)}  & 30.4 & 0.40 & 0.14 & 0.20 & 0.335 & 0.219\\
        {Qwen3.5-0.8B (\augmentation{}+)} & 30.5 & 0.29 & 0.05 & 0.08 & 0.339 & 0.183\\
        {Qwen3.5-2B (\augmentation{}+)} & {\bfseries 33.0} & 0.35 & {\bfseries 0.23} & {\bfseries 0.27} & 0.461 & 0.292\\
        \midrule
        \multicolumn{7}{c}{{Full recipes}} \\
        \midrule
        ProAssist~\cite{ZhangDLMKDCM25} & 0.2 & 0.00 & 0.00 & 0.00 & 0.000 & 0.000\\
        {Qwen3.5-2B (QICD)} &  8.2 & {0.18} & 0.06 & 0.09 & 0.347 & 0.161 \\
        {Qwen3.5-2B (\augmentation{}+)} & {\bfseries 14.8} & {\bfseries 0.24} & {\bfseries 0.12} & {\bfseries  0.16} & 0.446 & 0.276\\
        \bottomrule
    \end{tabularx}
\end{table}

\section{Experiments}
In this section, we first evaluate state of the art video LLMs on our \benchmark{} benchmark and highlight challenges in live intervention in video streams. 
Subsequently, we finetune on synthetic counterfactual data from \augmentation{} and show that it improves performance on \benchmark{}. 

\myparagraph{Evaluation metrics.} We metrics based on QICD~\cite{qicd_2025}: instruction completion accuracy (IC-Acc), mistake intervention (precision, recall, F1), and fluency (BERTScore, ROUGE-L). IC-Acc measures whether the model correctly identifies completed instructions; mistake intervention measures temporal alignment with ground-truth feedback; and fluency measures similarity to the reference feedback. Fluency is only directly comparable across models with similar mistake-intervention performance.

\begin{figure}[!t]
  \centering
  \includegraphics[width=0.85\linewidth]{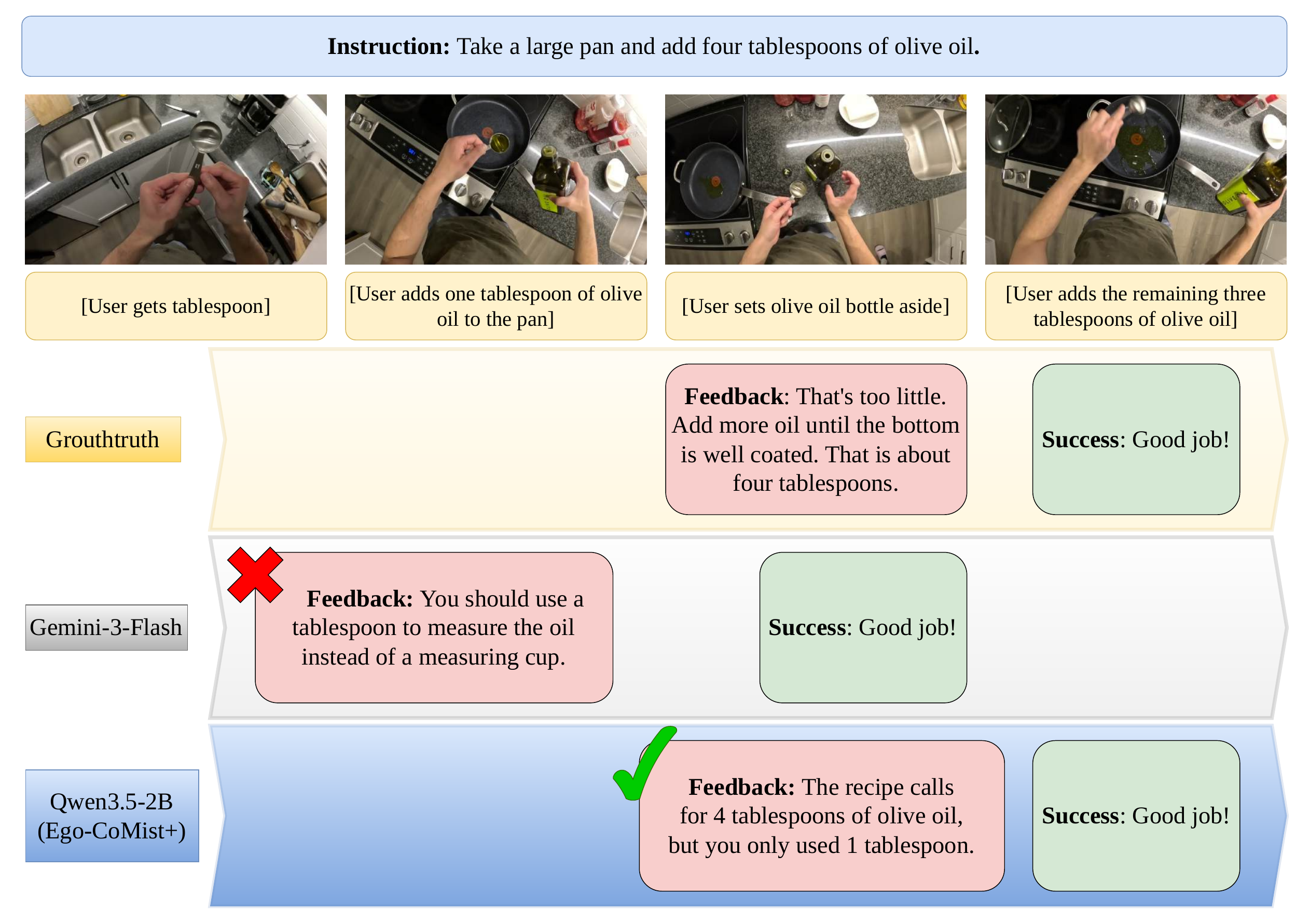}
  \includegraphics[width=0.85\linewidth]{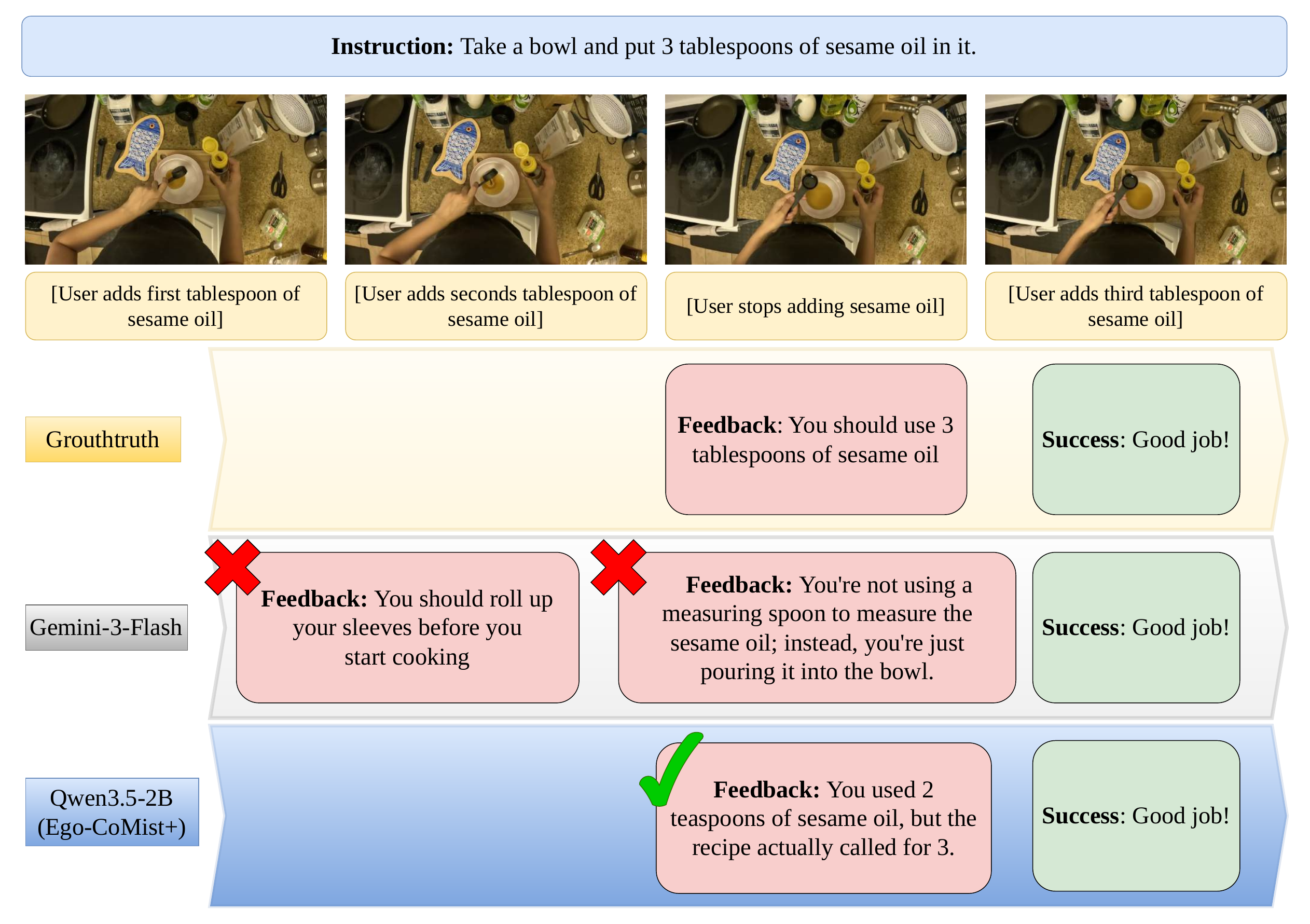}
  \vspace{-0.35cm}
  \caption{\benchmark{} streaming interventions: Gemini-3-Flash~\cite{gemini-flash} produces incorrect feedbacks and is unable to intervene when the person adds only one tablespoon of olive oil (top) and when the person adds only two tablespoons of sesame oil (bottom). The Qwen3.5-2B model finetuned on \augmentation{}+ intervenes at the appropriate time.}
  \label{fig:qualitative_ex1}
\end{figure}

\myparagraph{Evaluation protocol.}
\label{sec:streamcook_eval}
We consider two evaluation protocols: (i) \emph{per-recipe step}, (ii) 
\emph{full recipe step}~\cite{qicd_2025}. In the \emph{per-recipe step} evaluation setup, the model is provided with an instruction corresponding to a single recipe step, \eg, \emph{Take a large pan and add four tablespoons of olive oil.} as shown in \cref{fig:qualitative_ex1}, and the corresponding streaming video starting at the point where the instruction was provided in the groundtruth data. 
The task now is to provide appropriate feedbacks until the person successfully completes the recipe step. For the next recipe step, the input streaming video is re-started from the groundtruth recipe step start timestamp.
In the \emph{full recipe} protocol, the video is streamed continuously.
However, this setup  assumes that the user would move on to the next recipe step even when the model fails to generate a success confirmation message for the current step, which is unlikely in a real-world reactive setting (unlike the non-reactive setup of the Qualcomm Interactive Cooking Dataset~\cite{qicd_2025}).

\myparagraph{Evaluation of ``turn-based'' models.}
We begin by evaluating state of the art ``turn-based'' video LLMs on our \benchmark{} (\cref{tab:streamcook_zeroshot}): VideoLLaMA3-7B~\cite{damonlpsg2025videollama3}, Qwen2.5‑VL‑Instruct~\cite{abs-2502-13923}, Qwen3‑VL‑Instruct~\cite{abs-2511-21631}, Qwen3.5~\cite{qwen3p5} and Gemini‑Flash~\cite{gemini-flash,team2024gemini}. These are not streaming models and they only respond when prompted. 
We therefore adopt the regular‑interval prompting protocol~\cite{qicd_2025} for streaming evaluation, querying step completion and mistake intervention at fixed time intervals (a similar prompting strategy~\cite{shen2026simple} has recently shown state of the art performance on streaming QA benchmarks~\cite{li2025ovo,abs-2411-03628}).
We also consider the narration models:  {Videollm‑online}~\cite{VideoLLM-online} and {LiveCC}~\cite{chen2025livecc}. We convert their generated narrations into interactive feedback by passing them to a helper LLM (Qwen3-32B~\cite{abs-2505-09388}) that produces timely intervention messages (see appendix).


The results in \cref{tab:streamcook_zeroshot}, highlight that our \benchmark{} is highly challenging even for state of the art proprietary video LLMs such as Gemini-3-Flash. The mistake intervention F1 score remains low at 0.18 in the per-recipe step setting and importantly, the provided feedbacks are of limited utility as they do not match the ground  truth feedbacks closely as evidenced by the BERT and ROUGE-L metrics. This is further highlighted in \cref{fig:qualitative_ex1} (top), which shows the feedbacks provided by Gemini-3-Flash given the instruction: \emph{Take a large pan and add four tablespoons of olive oil.} The first feedback by Gemini-3-Flash asks the user to use a tablespoon to measure not a measuring cup, but the person has just taken out a tablespoon to measure. Afterwards, when the person sets the bottle of olive oil aside after adding just a single tablespoon of olive oil, Gemini-3-Flash assumes that the person has already completed the provided instruction.
Of the open-weights models in \cref{tab:streamcook_zeroshot}, the Qwen3.5 series of models performs best especially in the mistake intervention and fluency metrics. However, some models such as InternVL3.5-38B, Qwen2.5-VL-32B, Qwen3-VL-8B-Instruct, VideoLLaMA3-7B show poor mistake intervention performance. This is because they are unable to follow the mistake intervention instruction, likely because these models are designed for ``turn-based'' question-answering tasks (see appendix). 
The streaming Videollm-online and LiveCC models perform poorly as they do not produce informative narrations at the right time to be useful for intervention mistakes.
Finally, the performance in the full-recipe step setting is even weaker overall, with Gemini-3-Flash achieving the best mistake intervention performance and Qwen3.5-27B in the IC-Acc metric.

\myparagraph{Evaluation of ``streaming'' video LLMs.} 
Next, we evaluate streaming models that can provide interactive responses. To this end, we use our \augmentation{} dataset to fine-tune the Qwen3.5-2B, Qwen3.5-0.8B and Qwen3-VL-2B-Instruct~\cite{abs-2511-21631} models. 
To enable interactive responses using the ``turn-based'' Qwen models, we attach an action head on the last transformer block of these models. The action head predicts a binary speak/stay silent action after every input frame (\ie, after the \verb!<|vision_end|>! token)~\cite{qicd_2025,VideoLLM-online,panchal2024saysayitlive}. 
We also compare to the state of the art ProAssist~\cite{ZhangDLMKDCM25} model which is trained on conversational data in the cooking domain.

We report the baselines in \cref{tab:cc_finetune_streamcook}. The ProAssist model struggles to identify successful completion of instructions and provide fluent feedbacks.
The Qwen3.5-2B model fine-tuned on our \augmentation{} dataset performs better on all metrics.
To highlight the effectiveness of our \augmentation{} dataset, we consider an ablation where the Qwen3.5-2B model is finetuned instead on the QICD~\cite{qicd_2025} dataset. We see significant degradation of mistake intervention performance. 
This is to be expected as in QICD feedbacks are provided post-error and the participants are non-reactive.
Interestingly, we find that mixing QICD together with our \augmentation{} dataset (\augmentation{}+ in \cref{tab:cc_finetune_streamcook}) improves performance.  
This occurs only when we explicitly specify the qualitative difference in intervention strategies in the model prompt, leading to a multi-task learning setup.
The Qwen3.5-2B and Qwen3-VL-2B models trained on \augmentation{}+ outperform both ProAssist and Qwen3.5-2B (trained only on QICD), again highlighting the effectiveness of our \augmentation{} dataset.
Note that, the Qwen3.5-2B without fine-tuning on \augmentation{}+ performs very poorly as shown in \cref{tab:streamcook_zeroshot}.
Finetuning on \augmentation{}+ significantly improves its IC-Acc from 0.0 to 33.0 and mistake F1 from 0.04 to 0.18 and enables the model to react interactively to the input video stream -- matching performance of the proprietary Gemini-3-Flash model.

In the \emph{full recipe} evaluation setup, the performance of ProAssist is very weak: this is because as in the per-recipe step scenario, the model is unable to predict successful completion of recipe steps in the reactive setup of \benchmark{}. Moreover, unlike in the per-recipe step setup in the full-recipe setup errors accumulate over recipe steps, leading to poor performance. 
Finally, we again see that training on our \augmentation{} dataset leads to significant improvement over training just on QICD.

\section{Conclusion}
We introduced Qualcomm Interactive Cooking: \benchmark{}, a benchmark that addresses a key gap in interactive step‑by‑step task guidance: the evaluation of video LLMs in \emph{reactive} assistant–user interactions. 
We further proposed Qualcomm Interactive Cooking: \augmentation{}, a counterfactual synthetic dataset by converting non‑interactive cooking clips into supervised instances of proactive intervention, providing both counterfactual instructions and precisely timed feedback signals. Our evaluation shows that \benchmark{} is challenging for state‑of‑the‑art MLLMs: off‑the‑shelf models struggle to detect mistakes and to time interventions correctly. Fine‑tuning with \augmentation{} consistently improves mistake detection, intervention timing, and instruction‑completion accuracy.

\myparagraph{Limitations and Broader Impacts.} Our work is focused on the cooking domain through \benchmark{} and \augmentation{}. 
While we show that fine-tuning models on \augmentation{} leads to improved performance, these models are still far away from real-world deployment. Also, video LLMs can produce harmful and biased content, make incorrect claims and produce wrongful advice. This needs to be taken into account when interacting with, deploying or building on these models. It also has to be taken into account that any computer vision model processing visual information about human activities could in principle extract information beyond what is required for the use-case.

\myparagraph{Acknowledgment.} Shweta Mahajan was funded, in part, thanks to funding from the Canada Research Chairs Program and by the Natural Sciences and Engineering Research Council of Canada (NSERC) Discovery Grant.

\bibliographystyle{plainnat}
\bibliography{neurips_2026}

\newpage
\appendix

\section{Appendix}
Here we provide: \begin{enumerate*}
    \item Additional examples from our \benchmark{} benchmark.
    \item Additional details of the evaluation of ``turn-based'' video LLMs.
    \item Additional details of the evaluation of streaming narration models.
    \item Additional details of the evaluation metrics.
    \item Additional training details.
    \item Additional details of the \augmentation{} synthetic data generation pipeline.
    \item Prompts used at both stages of the \augmentation{} synthetic data generation process.
\end{enumerate*}

\section{\benchmark{}: Additional Examples}
\begin{figure}[t!]
    \centering
    \includegraphics[width=0.99\linewidth]{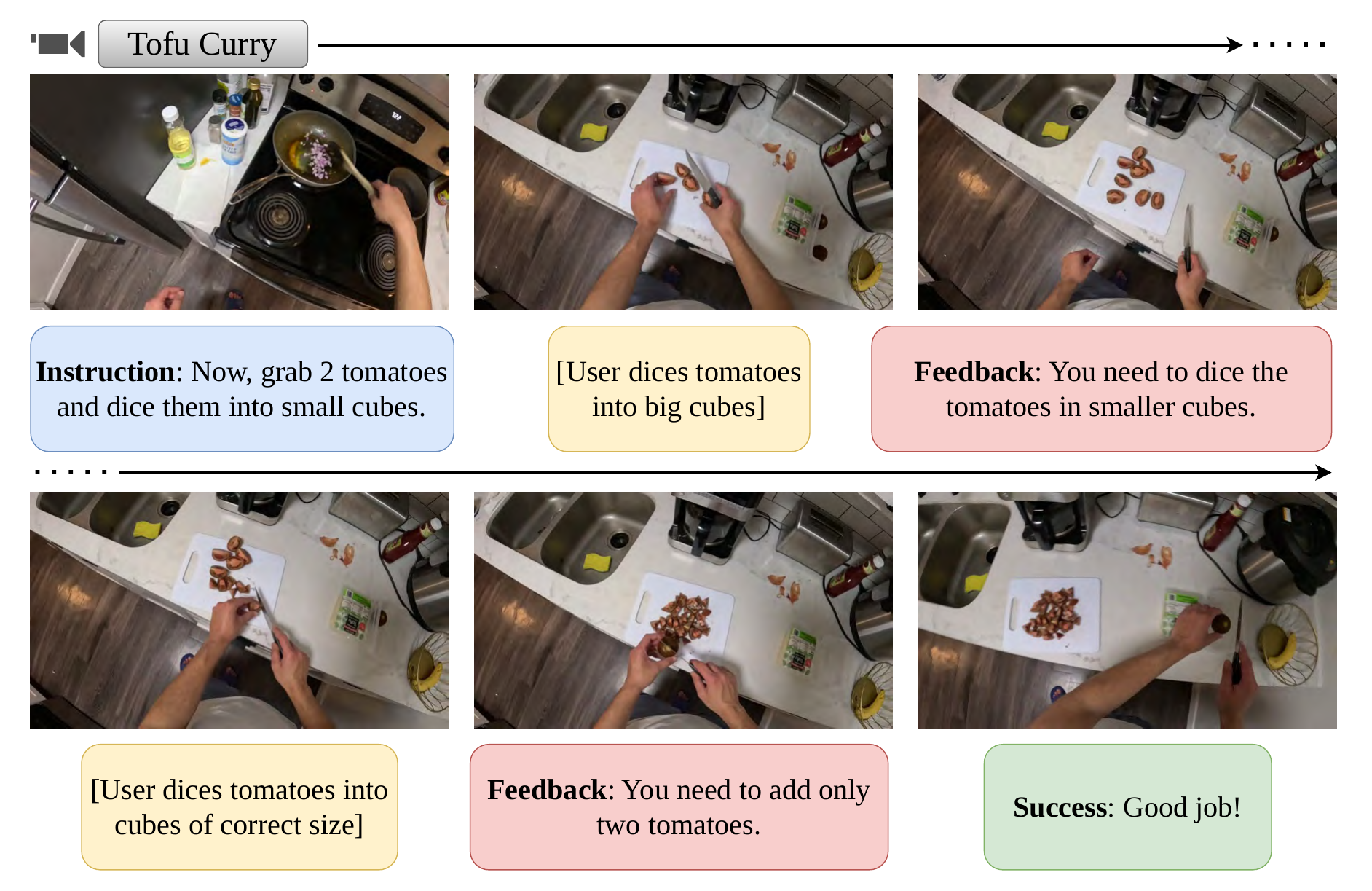}\\[0.25em]
    \includegraphics[width=0.99\linewidth]{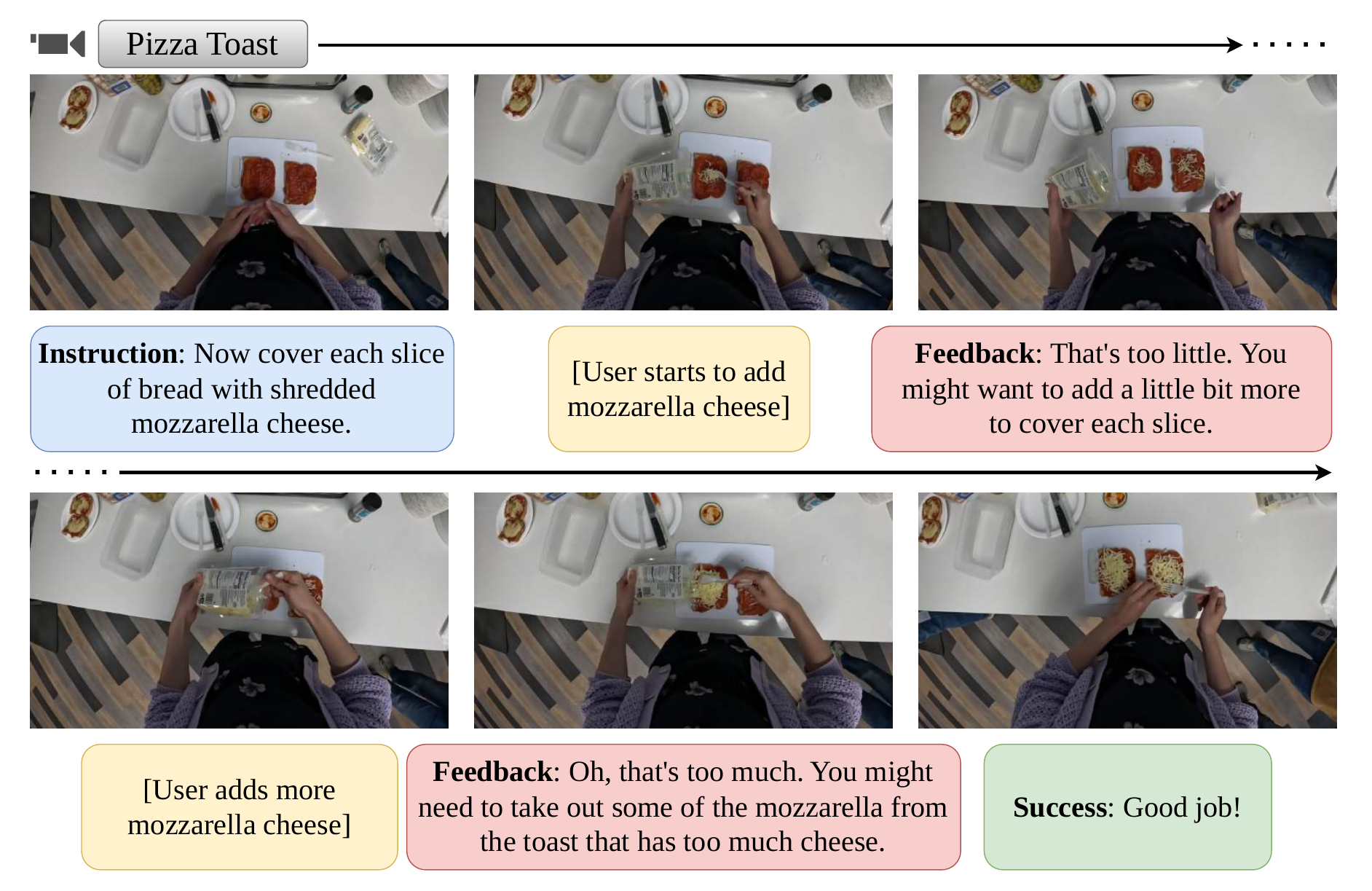}
    \caption{Our \benchmark{} benchmark includes interventions with appropriate feedback whenever a mistake is detected, guiding the user towards successful goal completion across recipe steps.}
    \label{fig:benchmark_examples_supp}
\end{figure}

We provide additional qualitative examples from our \benchmark{} benchmark in \cref{fig:benchmark_examples_supp}. 

In the top row, the assistant first provides the instruction: \emph{Now, grab 2 tomatoes
and dice them into small cubes}. The user then starts to dice the tomatoes, but make large cubes. So, the assistant intervenes and provides the feedback: \emph{You need to dice the
tomatoes in smaller cubes}. The user then dices the tomatoes correctly. But, instead of adding just two tomatoes, the user starts to dice a third tomato. The assistant intervenes again and provides the feedback: \emph{You need to add only two tomatoes}. The user then puts the tomato back and the assistant acknowledges the successful completion of the recipe step.

In the bottom row, the assistant first provides the instruction: \emph{Now cover each slice of bread with shredded mozzarella cheese}. The user then starts to add shredded mozzarella cheese to the slices of bread. But the user stops after adding too little cheese. So, the assistant intervenes and provides the feedback: \emph{That's too little. You might want to add a little bit more to cover each slice}. The user then adds way too much cheese on one of the slices. The assistant intervenes again and provides the feedback: \emph{Oh, that's too much. You might need to take out some of the mozzarella from the toast that has too much cheese}. The user then re-distributes the cheese correctly and the assistant acknowledges the successful completion of the recipe step.

These examples again highlight the core challenge of our \benchmark{} benchmark: intervention by providing appropriate feedback as soon as a mistake becomes apparent, guiding the user towards successful completion. 

\section{Additional Details: Evaluation of ``Turn-Based'' Models}
\label{sec:turn_based_eval}
As mentioned in the main paper, we use the regular interval prompting strategy~\cite{qicd_2025} to evaluate state of the art ``turn-based'' video LLMs which cannot respond interactively to video input streams: VideoLLaMA3-7B~\cite{damonlpsg2025videollama3}, InternVL3.5-38B~\cite{wang2025internvl3_5}, Qwen2.5‑VL‑Instruct~\cite{abs-2502-13923} series, Qwen3‑VL‑Instruct~\cite{abs-2511-21631} series, Qwen3.5~\cite{qwen3p5}  series, and Gemini‑Flash~\cite{team2024gemini,gemini-flash}. We use a time-interval of 5 seconds~\cite{qicd_2025} to balance accuracy and inference speed. At every turn we first prompt the video LLM to check if the person has completed the current recipe step:

\begin{tcolorbox}[
    colback=orange!5!white, 
    colframe=orange!75!black, 
    title=\textbf{Check if Recipe Step Complete},
    fonttitle=\bfseries,
    coltitle=black, 
    colbacktitle=orange!20!white, 
    sharp corners,
    boxrule=0.5pt
]
You are an expert cooking assistant who is observing a person cook.

\hrulefill 

\textbf{\#\#INSTRUCTIONS:}\\

The person is currently at the following recipe step: [recipe\_step]. Has the person already completed the recipe step?  If the person has completed the recipe step answer `YES' else answer `NO'. If you answer `YES' describe why you think the person already completed the recipe step.
\end{tcolorbox}

We found that asking the model to articulate why the recipe step is complete led to a boost in performance across models. Furthermore, we did not find any significant change in performance for alternate wordings of the prompt across models.

Now, if the model answers that the recipe step is not yet complete, we next ask the model to check if the person has made a mistake. For Qwen3‑VL‑Instruct~\cite{abs-2511-21631}, Qwen3.5~\cite{qwen3p5} and Gemini‑Flash~\cite{team2024gemini,gemini-flash} series, we use the following prompt which describes all mistake types in detail and also describes how to detect and intervene these mistakes. Also, in order to make sure that the video LLM does not repeat feedback for the same mistake again and again (as the past video frames are available in the context window), we store recently generated feedbacks (of the last 1 minute) in an array and ask the video LLM not to repeat the same feedback. Note that, this prompt was designed iteratively, where we added details until we saw no further improvements in performance.

\begin{tcolorbox}[
    colback=orange!5!white, 
    colframe=orange!75!black, 
    title=\textbf{Check for Mistake},
    fonttitle=\bfseries,
    coltitle=black, 
    colbacktitle=orange!20!white, 
    sharp corners,
    boxrule=0.5pt
]
You are an expert cooking assistant who is observing a person cook. You should look out for mistakes made by the person.

\hrulefill 

\textbf{\#\#INSTRUCTIONS:}\\
The person is trying to complete the following recipe step: [recipe\_step]. Your task is to check if the person is about to make or has already made a mistake. Mistakes occur when the person performs actions that deviates from the instruction and DIRECTLY INTERFERES WITH SUCCESSFUL INSTRUCTION COMPLETION. Do not penalize actions that do not directly interfere with instruction completion (e.g. washing broccoli before cutting, if the instruction just says `cut broccoli').

Here are some common types of mistakes that you should look out for:
1. Technique Error: A mistake in how a step is physically performed. Examples include chopping with the wrong motion, stirring when folding is required, or spilling during transfer, producing uneven cuts or texture issues even when tools and amounts are right. Ignore minor technique errors such as not holding or gripping objects properly, holding with a risk of dropping etc. that do not interfere directly with recipe completion. 

2. Preparation Error: A setup mistake before executing the step . Using the wrong or dirty utensil, not washing/peeling/draining ingredients, insufficient draining of fluid, cutting/chopping without peeling which makes correct execution difficult or unsafe.

3. Measurement Error: An error in quantity — wrong counts, volumes, weights, or units. Mixing up teaspoons and tablespoons, misreading a scale, or miscounting items leads to off ratios and predictable taste or texture problems.

4. Temperature Error: A mistake in heat level or thermal state — the applied temperature, starting temperature, or thermal transition is wrong. Not preheating, using the wrong microwave power, overheating oil, or adding cold liquid when warm is required often causes burning, undercooking, or split emulsions.

5. Timing Error: A mistake in duration -- over- or under-doing a step or skipping required rests, proofs, or cooling periods. Overcooking, underblending, or cutting resting time short typically yields incorrect doneness or unstable textures.\\

Do not repeat previously detected mistakes. Here are the feedbacks corresponding to previously detected mistakes:[prev\_detected\_mistakes]. Assume the recipe step is still in progress. Your task is to identify any mistake that's already visible in the partially completed step. Do no penalize partially competed recipe steps. If you observe a mistake answer `YES', else `No'. Your response MUST BEGIN WITH `YES' or `NO'. In case you answer `YES', please follow with a concise feedback to the user describing the mistake (i.e. YES. <feedback>.). Directly address the person.
\end{tcolorbox}

For the VideoLLaMA3-7B~\cite{damonlpsg2025videollama3}, InternVL3.5-38B~\cite{wang2025internvl3_5}, Qwen2.5‑VL‑Instruct~\cite{abs-2502-13923} series, we used the following simple prompt:

\begin{tcolorbox}[
    colback=orange!5!white, 
    colframe=orange!75!black, 
    title=\textbf{Check for Mistake},
    fonttitle=\bfseries,
    coltitle=black, 
    colbacktitle=orange!20!white, 
    sharp corners,
    boxrule=0.5pt
]
You are an expert cooking assistant who is observing a person cook. You should look out for mistakes made by the person.

\hrulefill 

\textbf{\#\#INSTRUCTIONS:}\\
The person is trying to complete the following recipe step: [recipe\_step]. Your task is to check if the person is about to make or has already made a mistake. Mistakes occur when the person performs actions that deviates from the instruction and DIRECTLY INTERFERES WITH SUCCESSFUL INSTRUCTION COMPLETION. Assume the recipe step is still in progress. Your task is to identify any mistake that's already visible in the partially completed step. \\

Do no penalize partially competed recipe steps. If you observe a mistake answer `YES', else `No'. Your response MUST BEGIN WITH `YES' or `NO'. In case you answer `YES', please follow with a concise feedback to the user describing the mistake (i.e. YES. <feedback>.). Directly address the person.
\end{tcolorbox}

As shown in \cref{tab:streamcook_zeroshot}, even with this simple prompt these models (almost) never predict that a mistake has occurred. Adding the additional details as in the prompt for the Qwen3‑VL‑Instruct~\cite{abs-2511-21631}, and Gemini‑Flash~\cite{gemini-flash,team2024gemini} series does not improve performance. 

\section{Additional Details: Evaluation of Streaming Narration Models}
As mentioned in the main paper, for streaming narration models such as Videollm-online~\cite{VideoLLM-online} and LiveCC~\cite{chen2025livecc}, we convert their generated narrations into interactive feedback by passing them to a helper LLM (Qwen3.5-27B~\cite{qwen3p5}). Note that in case of LiveCC, it produces narrations after every 1 second of video. We store these narrations in a buffer ([narration\_list]) and after every 5 seconds (similar to the case of the ``turn-based'' video LLMs described above in \cref{sec:turn_based_eval}), we prompt the helper Qwen3.3-27B~\cite{qwen3p5} LLM as follows:
\begin{tcolorbox}[
    colback=orange!5!white, 
    colframe=orange!75!black, 
    title=\textbf{Check if Recipe Step Complete (Streaming Narration Models)},
    fonttitle=\bfseries,
    coltitle=black, 
    colbacktitle=orange!20!white, 
    sharp corners,
    boxrule=0.5pt
]
You are an intelligent chatbot that is judging another system which narrates human cooking videos.  Given a high level action instruction and a list of narrations generated from the system, your job is to decide if the narration is correct and shows completion of the instruction. 

\hrulefill 

\textbf{\#\#INSTRUCTIONS:}\\
List of narrations: [narration\_list] \\

Answer `YES' if the instruction is completed otherwise output `NO'.
\end{tcolorbox}

If the helper LLM predicts that the instruction is not complete, we next ask the helper LLM to check if the person has made a mistake (similar to the case of the ``turn-based'' video LLMs). We use the simple prompt described in \cref{sec:turn_based_eval} used with the VideoLLaMA3-7B~\cite{damonlpsg2025videollama3}, InternVL3.5-38B~\cite{wang2025internvl3_5}, Qwen2.5‑VL‑Instruct~\cite{abs-2502-13923} series, Qwen3.5~\cite{qwen3p5} series models. But instead of the video we provide the narrations from the streaming narration model as input. Adding additional details to the prompt as for the Qwen3‑VL‑Instruct/Qwen3.5, and Gemini‑Flash series does not improve performance. 

\section{Additional Details: Evaluation Metrics}
To compute the IC-Acc, mistake intervention and mistake fluency metrics, we use a temporal window size of 30 seconds~\cite{qicd_2025}. 

The mistake fluency scores are comparable only at the same mistake detection levels, because the scores are computed only for true positives. If the true positive rate is low, then the model could get higher scores by making a few very good predictions. But in practice, a model with a higher true positive rate is preferred.

\section{Additional Training Details}
As described in the main paper, we fine-tune the language backbone of the Qwen3-VL-2B-Instruct, Qwen3.5-0.8B, Qwen3.5-2B models using LoRA (dim = 64) for $\sim$100k iterations (based on validation loss). We use the AdamW optimizer with a learning rate of $2\times 10^{-4}$ and a cosine annealing learning rate schedule for 100k iterations until a learning rate of $1\times 10^{-6}$. We use a batch size of 32 using 8 Nvidia H100 GPUs (with 4 gradient accumulation steps).
For all Qwen3-VL/3.5 models in \cref{tab:cc_finetune_streamcook} trained using \augmentation{}+ data, we additionally train them on action segments from Ego4D Goal-Step~\cite{SongBNWMT23} to improve IC-Acc scores. We convert the leaf action descriptions in Ego4D Goal-Step into instructions (using Qwen3-8B) and add success confirmation messages at the end of the action.
Overall composition of \augmentation{}+ is 60\% \augmentation{}, 30\% QICD, and 10\% Ego4D Goal-Step actions.

\section{\augmentation{}: Data Generation Process}
\subsection{Stage 1: Counterfactual Instruction and Feedback Generation}
Continuing from \cref{sec:stage_1_ego_comist} in the main paper, we describe the stage 1 of the counterfactual data generation pipeline for the preparation, technique, temperature and timing errors. 
\myparagraph{Preparation error.} Given a recipe step which uses a specific preparation method on an ingredient or item (object) used in the cooking process, we first ask the LLM to extract the preparation method and the object. 
For example, given the instruction: \emph{Coat the 6 oz. ramekin cup with cooking spray}, the preparation method is \emph{coat with cooking spray} and the object is the \emph{ramekin cup}.
Then, we ask the LLM to propose an alternative preparation method that for the object that aligns with the original recipe step.
For example: \emph{coat with cooking spray} can be replaced with \emph{grease with butter}.
Given the original preparation method and the alternative preparation method, we can now generate the counterfactual instruction and feedback messages: \emph{Grease the 6 oz. ramekin cup with butter} and \emph{You should grease the ramekin cup with butter, not coat with cooking spray}.

\myparagraph{Technique error.} Similar to the counterfactual instruction and feedback generation process for preparation errors described above, given a recipe step, we extract the specific cooking technique used along with the ingredient or item (object) used in the specific technique.
For example, given the instruction: \emph{Cut the English muffin into two pieces with a knife}, the technique is \emph{cut with a knife} and the object is the \emph{English muffin}.
Then, we ask the LLM to propose an alternative technique that aligns with the original recipe step.
For example: \emph{cut with a knife} can be replaced with \emph{split with a fork}.
Given, the original and the alternative technique, we can now generate the counterfactual instruction and feedback messages: \emph{Split the English muffin into two pieces with a fork} and \emph{Don't use a knife to cut the muffin, use a fork to split it instead}.

\myparagraph{Temperature error.} Given a recipe step that involves cooking at a specific temperature, heat setting or object at a thermal state (\eg, hot/cold), we ask the LLM to suggest an alternative reasonable temperature, heat setting or a thermal state. For example, given the recipe step \emph{Microwave the plate, covered, on high for 1.5 minutes}, we can generate the counterfactual instruction: \emph{Microwave the plate, covered, at medium heat for 1.5 minutes} and the feedback: \emph{Be careful, this step calls for medium heat, not high, to avoid overcooking}. 

\myparagraph{Timing error.} Given a recipe step that involves cooking for a specific duration, we ask the LLM to suggest an alternative reasonable duration. For example, given the recipe step \emph{Microwave on high for 30 seconds}, we can generate the counterfactual instruction: \emph{Microwave on high for 45 seconds} and the feedback: \emph{You did not microwave for long enough. You should microwave on high for 45 seconds}. Additionally, to keep the dataset balanced, we generate an equal number of counterfactual instructions with shorter and longer durations.

\subsection{Stage 2: Counterfactual Feedback Timestamp Inference}
Continuing from \cref{sec:stage_2_ego_comist} in the main paper, we describe the stage 2 of the counterfactual data generation pipeline for the preparation, technique, temperature and timing errors. 

\myparagraph{Preparation error.} In case of preparation errors, we prompt the LLM to find the timestamp at which the person starts to use the incorrect preparation method. For example, if the original instruction was \emph{Coat the 6 oz. ramekin cup with cooking spray} and the counterfactual instruction is \emph{Grease the 6 oz. ramekin cup with butter}, we prompt the LLM to look for the description which states that the person is starting to coat the ramekin cup with cooking spray. We use the timestamp of this description as the counterfactual feedback timestamp.

\myparagraph{Technique error.} In case of technique errors, we prompt the LLM to find the timestamp where the person starts to use the incorrect technique. For example, if the original instruction was \emph{Cut the English muffin into two pieces with a knife} and the counterfactual instruction is \emph{Split the English muffin into two pieces with a fork}, we prompt the LLM to find the description from the step by step descriptions which states that the person starts to cut the english muffin into two pieces with a knife. We use the timestamp of this description as the counterfactual feedback timestamp.

\myparagraph{Temperature error.} In case of temperature errors, we prompt the LLM to find the timestamp where it becomes clear that the person has used the wrong heat setting. For example, if the original instruction calls for microwaving at high heat and the counterfactual instruction calls for microwaving at medium heat, we use the timestamp where the person sets the heat setting on the microwave.

\myparagraph{Timing error.} In case the counterfactual amount of time ($y$\,seconds) is smaller than the original amount of time ($x$\,seconds) in the recipe step, we provide the feedback at least $y$\,seconds from when the cooking process begins. For example, if the original recipe calls for microwaving for 45\,seconds and the counterfactual amount of time is 30\,seconds, we wait at least 30\,seconds from the beginning of the microwaving process. In case the counterfactual amount of time is larger than the original amount of time, we provide the feedback after the cooking process finishes, \ie, after $x$\,seconds from the beginning of the cooking process.

\section{\augmentation{} Prompts: Stage 1}
Here we provide the prompts used in stage 1 (Counterfactual Instruction and Feedback Generation) of the \augmentation{} data generation pipeline.

\subsection{Measurement error}
As described in the main paper, given a recipe step: [org\_recipe\_step], we first ask the LLM to extract the attributes: quantity [quantity] and the corresponding ingredient [ingredient].

\begin{tcolorbox}[
    colback=blue!5!white, 
    colframe=blue!75!black, 
    title=\textbf{Extract Attributes from Recipe Step},
    fonttitle=\bfseries,
    coltitle=black, 
    colbacktitle=blue!20!white, 
    sharp corners,
    boxrule=0.5pt
]
You are an expert cooking assistant. You are watching a person follow a step by step recipe.

\hrulefill 

\textbf{\#\#INSTRUCTIONS:}\\
The person is following the recipe step: [org\_recipe\_step].  \\

Does this recipe step involve measuring a specific quantity or amount of an ingredient? \\

Examples include specific quantities like teaspoon, teaspoons, tablespoons, liters, ounces, grams, kgs, kilograms, cups, pints, quarts, gallons etc. \\

Extract the explicitly stated numerical quantity or amount and Answer `YES' or `NO'. If `YES', then return the specific quantity and  ingredient. Return a dict with two fields: \{`ingredient': .., `quantity': ...\}. DO NOT RETURN ANYTHING ELSE.
\end{tcolorbox}

Then, we ask the LLM to create a counterfactual quantity [counterfactual\_quantity], reasonable for the given recipe step. We use different prompts for greater and smaller quantities, which ask the LLM to generate greater or smaller counterfactual quantities respectively, as shown below.

\begin{tcolorbox}[
    colback=blue!5!white, 
    colframe=blue!75!black, 
    title=\textbf{Counterfactual Attributes (Greater)},
    fonttitle=\bfseries,
    coltitle=black, 
    colbacktitle=blue!20!white, 
    sharp corners,
    boxrule=0.5pt
]
You are an expert cooking assistant.

\hrulefill 

\textbf{\#\#INSTRUCTIONS:}\\
The following recipe step: [org\_recipe\_step]; uses the following quantity: [quantity] of the item: [ingredient]. 
Suggest an alternative reasonable quantity of the item greater than [quantity] for the recipe step. 
RETURN JUST THE QUANTITY. DO NOT RETURN ANYTHING ELSE.
\end{tcolorbox}

Based on the original and counterfactual quantities, we generate the counterfactual instruction and feedback pair.

\begin{tcolorbox}[
    colback=blue!5!white, 
    colframe=blue!75!black, 
    title=\textbf{Counterfactual Instruction},
    fonttitle=\bfseries,
    coltitle=black, 
    colbacktitle=blue!20!white, 
    sharp corners,
    boxrule=0.5pt
]
You are an expert cooking assistant.

\hrulefill 

\textbf{\#\#INSTRUCTIONS:}\\
Modify the instruction [org\_instruction] for the recipe step: [org\_recipe\_step]; such that the recipe step uses [counterfactual\_quantity] of [ingredient]. 
RETURN JUST THE INSTRUCTION DO NOT RETURN ANYTHING ELSE.
\end{tcolorbox}

\begin{tcolorbox}[
    colback=blue!5!white, 
    colframe=blue!75!black, 
    title=\textbf{Counterfactual Feedback},
    fonttitle=\bfseries,
    coltitle=black, 
    colbacktitle=blue!20!white, 
    sharp corners,
    boxrule=0.5pt
]
You are an expert cooking assistant. You are assisting a person step by step through a recipe.

\hrulefill 

\textbf{\#\#INSTRUCTIONS:}\\
Instead of using the specified quantity: [quantity] of [ingredient], the person instead used the following quantity: [counterfactual\_quantity]. Given the correct specified quantity and the incorrectly used quantity provide an appropriate feedback message. The feedback message should be short and one line in length. Make sure to point out the mistake clearly in the feedback message in a single line. RETURN JUST THE FEEDBACK MESSAGE DO NOT RETURN ANYTHING ELSE.
\end{tcolorbox}

\subsection{Preparation error}
As described in the main paper, given a recipe step  [org\_recipe\_step], we ask the LLM to extract the preparation method [current\_prep\_method] and the object of the preparation method [object] and then propose a counterfactual (alternative) preparation method [alternative\_prep\_method] for the object that aligns with the original recipe step. Note, that in the prompt we use ``alternative'' instead of ``counterfactual'' for clarity.

\begin{tcolorbox}[
    colback=blue!5!white, 
    colframe=blue!75!black, 
    title=\textbf{Extract Attributes and Propose Counterfactual Attributes},
    fonttitle=\bfseries,
    coltitle=black, 
    colbacktitle=blue!20!white, 
    sharp corners,
    boxrule=0.5pt
]
You are an expert cooking assistant. 

\hrulefill 

\textbf{\#\#INSTRUCTIONS:}\\
Given a recipe step, your task is to propose variants of that recipe step that use an alternative preparation method. Alternative preparation methods include: different blending/whisking/mixing/beating methods; using different utensils or ingredients; using left instead of right hands and vise versa; dealing with fluids differently; cutting or chopping ingredients differently.  \\

Now, given the following recipe step: [org\_recipe\_step]. Extract the specific preparation method used, the food item or cooking utensil used (object) and then propose an alternative preparation method. Make sure that the proposed alternative preparation method is realistic and likely to be provided by a expert cooking assistant. \\

Return a dict with five keys: \{`current\_prep\_method': ..., `object': ... , `alternative\_prep\_method': ... , `current\_prep\_method\_duration': ..., `alternative\_prep\_method\_duration': ...\}. Where, if the current preparation method (current\_prep\_method) requires cooking (e.g. heating, boiling, microwaving) for a specific duration this duration is returned in [current\_prep\_method\_duration] field (None otherwise); and if the alternative preparation method (alternative\_prep\_method\_duration) requires cooking (e.g. heating, boiling, microwaving) for a specific duration this duration is returned in [alternative\_prep\_method\_duration] field (None otherwise). Keep the [alternative\_prep\_method] in the dict very short. RETURN JUST THE COOKING PREP METHOD DICT DO NOT RETURN ANYTHING ELSE.
\end{tcolorbox}

Based on the original and counterfactual preparation methods, we generate the counterfactual instruction and feedback pair.

\begin{tcolorbox}[
    colback=blue!5!white, 
    colframe=blue!75!black, 
    title=\textbf{Counterfactual Instruction},
    fonttitle=\bfseries,
    coltitle=black, 
    colbacktitle=blue!20!white, 
    sharp corners,
    boxrule=0.5pt
]
You are an expert cooking assistant.

\hrulefill 

\textbf{\#\#INSTRUCTIONS:}\\
\end{tcolorbox}

\begin{tcolorbox}[
    colback=blue!5!white, 
    colframe=blue!75!black, 
    fonttitle=\bfseries,
    coltitle=black, 
    colbacktitle=blue!20!white, 
    sharp corners,
    boxrule=0.5pt
]

Generate a cooking instruction that uses the following preparation method: [alternative\_prep\_method] on: [object] for: [alternative\_prep\_method\_duration]. Here is an example of a similar instruction: [org\_instruction]. RETURN JUST THE INSTRUCTION DO NOT RETURN ANYTHING ELSE. 
\end{tcolorbox}

\begin{tcolorbox}[
    colback=blue!5!white, 
    colframe=blue!75!black, 
    title=\textbf{Counterfactual Feedback},
    fonttitle=\bfseries,
    coltitle=black, 
    colbacktitle=blue!20!white, 
    sharp corners,
    boxrule=0.5pt
]
You are an expert cooking assistant.

\hrulefill 

\textbf{\#\#INSTRUCTIONS:}\\
Instead of the correct preparation method: [org\_prep\_method] the person used the preparation method: [alternative\_prep\_method] -- on the food item or cooking utensil: [object]. Provide an appropriate feedback message to the person. The feedback message should be one line in length and clearly point out the mistake made by the person. Make sure to point out the mistake clearly in the feedback message in a single line. RETURN JUST THE FEEDBACK MESSAGE DO NOT RETURN ANYTHING ELSE.
\end{tcolorbox}

\subsection{Technique error} 
As described in the main paper, given a recipe step  [org\_recipe\_step], we ask the LLM to extract the specific cooking technique used [technique] along with the ingredient or item (object) [item] used in the specific technique.

\begin{tcolorbox}[
    colback=blue!5!white, 
    colframe=blue!75!black, 
    title=\textbf{Extract Attributes from Recipe Step},
    fonttitle=\bfseries,
    coltitle=black, 
    colbacktitle=blue!20!white, 
    sharp corners,
    boxrule=0.5pt
]
You are an expert cooking assistant. You are assisting a person step by step through a recipe.

\hrulefill 

\textbf{\#\#INSTRUCTIONS:}\\
The person is trying to complete the following recipe step: [org\_recipe\_step].\\

If the recipe step uses a specific cooking technique, example techniques include: roll, pat, squeeze, hold, measuring, adding, transferring, chopping, cutting, peeling, spiralizing, flipping, stirring, whisking, beating, return the cooking technique, return a dict with the cooking technique and the food item or ingredient that it is used on: \{`technique': ..., `item':...\}.
If not, return None. DO NOT RETURN ANYTHING ELSE.
\end{tcolorbox}

Then, we ask the LLM to propose a counterfactual technique [counterfactual\_technique] that aligns with the original recipe step. In the prompt, again we use ``alternative'' instead of ``counterfactual'' for clarity.

\begin{tcolorbox}[
    colback=blue!5!white, 
    colframe=blue!75!black, 
    title=\textbf{Counterfactual Attributes},
    fonttitle=\bfseries,
    coltitle=black, 
    colbacktitle=blue!20!white, 
    sharp corners,
    boxrule=0.5pt
]
You are an expert cooking assistant.

\hrulefill 

\textbf{\#\#INSTRUCTIONS:}\\
Given a recipe step, [org\_recipe\_step] that uses the cooking technique: [technique] on [item], your task is to propose an alternative cooking technique. 
Make sure that the proposed alternative cooking technique is realistic given the recipe step. RETURN JUST THE ALTERNATIVE COOKING TECHNIQUE DO NOT RETURN ANYTHING ELSE. 
\end{tcolorbox}

Based on the original and counterfactual techniques, we generate the counterfactual instruction and feedback pair.

\begin{tcolorbox}[
    colback=blue!5!white, 
    colframe=blue!75!black, 
    title=\textbf{Counterfactual Instruction},
    fonttitle=\bfseries,
    coltitle=black, 
    colbacktitle=blue!20!white, 
    sharp corners,
    boxrule=0.5pt
]
You are an expert cooking assistant.

\hrulefill 

\textbf{\#\#INSTRUCTIONS:}\\
Given the cooking instruction [org\_instruction], generate a new instruction that uses the following cooking technique instead: [alternative\_technique]. RETURN JUST THE INSTRUCTION DO NOT RETURN ANYTHING ELSE. 
\end{tcolorbox}

\begin{tcolorbox}[
    colback=blue!5!white, 
    colframe=blue!75!black, 
    title=\textbf{Counterfactual Feedback},
    fonttitle=\bfseries,
    coltitle=black, 
    colbacktitle=blue!20!white, 
    sharp corners,
    boxrule=0.5pt
]
You are an expert cooking assistant.

\hrulefill 

\textbf{\#\#INSTRUCTIONS:}\\
Instead of the correct technique: [counterfactual\_technique] the person used the technique: [technique] -- on the: [item]. Provide an appropriate feedback message to the person. The feedback message should be one line in length and clearly point out the mistake made by the person. Make sure to point out the mistake clearly in the feedback message in a single line. RETURN JUST THE FEEDBACK MESSAGE DO NOT RETURN ANYTHING ELSE.
\end{tcolorbox}

\subsection{Temperature error}
As described in the main paper, given a recipe step  [org\_recipe\_step], we first ask the LLM to extract the cooking temperature or heat setting [heat\_setting] using the kitchen appliance or utensil [kitchen\_item] used on the [food\_item].

\begin{tcolorbox}[
    colback=blue!5!white, 
    colframe=blue!75!black, 
    title=\textbf{Extract Attributes from Recipe Step},
    fonttitle=\bfseries,
    coltitle=black, 
    colbacktitle=blue!20!white, 
    sharp corners,
    boxrule=0.5pt
]
You are an expert cooking assistant. You are assisting a person step by step through a recipe.

\hrulefill 

\textbf{\#\#INSTRUCTIONS:}\\
The person is following the recipe step: [org\_recipe\_step]\\

Does this recipe step involve cooking an ingredient or a food item on a kitchen appliance or utensil at a specific temperature or heat setting? Examples include setting an oven or a stove at a specific temperature or heat setting. If yes, then return the specific kitchen appliance or utensil (kitchen item), temperature or heat setting (heat setting) and the food item or ingredient (food item). Return a dict with three fields: \{`kitchen\_item': .., `heat\_setting': ..., `food\_item': ...\} else return None. DO NOT RETURN ANYTHING ELSE.
\end{tcolorbox}

Then, we ask the LLM to suggest a counterfactual temperature or a heat setting [counterfactual\_heat\_setting]. We use separate (similar) prompts for higher or lower heat settings. In the prompt, we again use ``alternative'' instead of ``counterfactual'' for clarity.

\begin{tcolorbox}[
    colback=blue!5!white, 
    colframe=blue!75!black, 
    title=\textbf{Counterfactual Attribute (Higher)},
    fonttitle=\bfseries,
    coltitle=black, 
    colbacktitle=blue!20!white, 
    sharp corners,
    boxrule=0.5pt
]
You are an expert cooking assistant. 

\hrulefill 

\textbf{\#\#INSTRUCTIONS:}\\
The following recipe step: [org\_recipe\_step]; uses the following kitchen appliance/utensil: [kitchen\_item]; to cook the following food item: [food\_item]; at the following heat setting: [heat\_setting]. Suggest an alternative reasonable heat setting higher than [heat\_setting] for the recipe step. RETURN JUST THE HEAT SETTING IF POSSIBLE ELSE RETURN NONE. DO NOT RETURN ANYTHING ELSE. 
\end{tcolorbox}

Based on the original and counterfactual heat settings, we generate the counterfactual instruction and feedback pair.

\begin{tcolorbox}[
    colback=blue!5!white, 
    colframe=blue!75!black, 
    title=\textbf{Counterfactual Instruction},
    fonttitle=\bfseries,
    coltitle=black, 
    colbacktitle=blue!20!white, 
    sharp corners,
    boxrule=0.5pt
]
You are an expert cooking assistant.

\hrulefill 

\textbf{\#\#INSTRUCTIONS:}\\
Modify the cooking instruction: [org\_instruction]; such that the recipe step uses the following heat setting: [counterfactual\_heat\_setting] -- while cooking [food\_item] on the utensil/appliance: [kitchen\_item]. RETURN JUST THE INSTRUCTION DO NOT RETURN ANYTHING ELSE. 
\end{tcolorbox}

\begin{tcolorbox}[
    colback=blue!5!white, 
    colframe=blue!75!black, 
    title=\textbf{Counterfactual Feedback},
    fonttitle=\bfseries,
    coltitle=black, 
    colbacktitle=blue!20!white, 
    sharp corners,
    boxrule=0.5pt
]
You are an expert cooking assistant.

\hrulefill 

\textbf{\#\#INSTRUCTIONS:}\\
Instead of using the correct heat setting: [heat\_setting], the person instead used a incorrect heat setting: [counterfactual\_heat\_setting] -- on the food item: [food\_item] -- and using the kitchen utensil/appliance: [kitchen\_item]. \\

Provide an appropriate feedback message. The feedback message should be short and one line in length. Make sure to point out the heat setting clearly in a feedback message in a single line. RETURN JUST THE FEEDBACK MESSAGE DO NOT RETURN ANYTHING ELSE. 
\end{tcolorbox}

\subsection{Timing error}
As described in the main paper, given a recipe step  [org\_recipe\_step], we first ask the LLM to extract the duration [amount\_of\_time], the food item or ingredient [food\_item] and the specific kitchen appliance or utensil [kitchen\_item] the [food\_item] is cooked in.

\begin{tcolorbox}[
    colback=blue!5!white, 
    colframe=blue!75!black, 
    title=\textbf{Extract Attributes from Recipe Step},
    fonttitle=\bfseries,
    coltitle=black, 
    colbacktitle=blue!20!white, 
    sharp corners,
    boxrule=0.5pt
]
You are an expert cooking assistant. You are assisting a person step by step through a recipe.

\hrulefill 

\textbf{\#\#INSTRUCTIONS:}\\
The person is following the recipe step: [org\_recipe\_step]. Does this recipe step involve cooking an ingredient or a food item on a kitchen appliance or utensil for a specific amount of time? Examples include setting an oven or a stove at a specific temperature or heat setting.\\ 

If yes, then return the specific kitchen appliance or utensil (kitchen\_item), the amount of time (amount\_of\_time) and the food item or ingredient (food\_item). Return a dict with three fields: \{`kitchen\_item': .., `amount\_of\_time': ..., `food\_item': ...\}. Else return None. DO NOT RETURN ANYTHING ELSE.
\end{tcolorbox}

Then, we ask the LLM to propose a counterfactual reasonable amount of time [counterfactual\_amount\_of\_time]. In the prompt, we again use ``alternative'' instead of ``counterfactual'' for clarity.

\begin{tcolorbox}[
    colback=blue!5!white, 
    colframe=blue!75!black, 
    title=\textbf{Counterfactual Attribute (Greater)},
    fonttitle=\bfseries,
    coltitle=black, 
    colbacktitle=blue!20!white, 
    sharp corners,
    boxrule=0.5pt
]
You are an expert cooking assistant. 

\hrulefill 

\textbf{\#\#INSTRUCTIONS:}\\
The following recipe step: [org\_recipe\_step]; uses the following kitchen appliance/utensil: [kitchen\_item]; to the cook the following food item: [food\_item]; for the following amount of time: [amount\_of\_time]. Suggest an alternative reasonable amount of time significantly greater (at least 15 seconds) than [amount\_of\_time] for the recipe step. RETURN JUST THE AMOUNT OF TIME IF POSSIBLE ELSE RETURN NONE. DO NOT RETURN ANYTHING ELSE.
\end{tcolorbox}

Based on the original and counterfactual amounts of time, we generate the counterfactual instruction and feedback pair.

\begin{tcolorbox}[
    colback=blue!5!white, 
    colframe=blue!75!black, 
    title=\textbf{Counterfactual Instruction},
    fonttitle=\bfseries,
    coltitle=black, 
    colbacktitle=blue!20!white, 
    sharp corners,
    boxrule=0.5pt
]
You are an expert cooking assistant.

\hrulefill 

\textbf{\#\#INSTRUCTIONS:}\\
Modify the cooking instruction: [org\_instruction]; such that the recipe step uses the following amount of time: [counterfactual\_amount\_of\_time] -- while cooking [food\_item] on the utensil/appliance: [kitchen\_item]. RETURN JUST THE INSTRUCTION DO NOT RETURN ANYTHING ELSE. 
\end{tcolorbox}

\begin{tcolorbox}[
    colback=blue!5!white, 
    colframe=blue!75!black, 
    title=\textbf{Counterfactual Feedback (Greater)},
    fonttitle=\bfseries,
    coltitle=black, 
    colbacktitle=blue!20!white, 
    sharp corners,
    boxrule=0.5pt
]
You are an expert cooking assistant. You are assisting a person step by step through a recipe.

\hrulefill 

\textbf{\#\#INSTRUCTIONS:}\\
Instead of the using the correct amount of time for cooking: [amount\_of\_time, the person did not cook the food item: [food\_item] -- and using the kitchen utensil/appliance: [kitchen\_item] -- for long enough. Provide an appropriate feedback message. The feedback message should be short and one line in length. Make sure to point out that the person did not cook for long enough clearly in the feedback message in a single line. RETURN JUST THE FEEDBACK MESSAGE DO NOT RETURN ANYTHING ELSE.
\end{tcolorbox}

\section{\augmentation{} Prompts: Stage 2}
Here we provide the prompts used in the stage 2 (Counterfactual Feedback Timestamp Inference) of the \augmentation{} data generation pipeline. The prompts always use the step by step descriptions [step\_by\_step\_descriptions] extracted using the state of the art video LLM at timestamps [timestamps] that are 5 seconds apart using a sliding window, as described in the main paper.

\subsection{Measurement error}
In addition to the step by step descriptions [step\_by\_step\_descriptions] at timestamps [timestamps], we use the attributes [quantity], [ingredient] and the counterfactual quantity [counterfactual\_quantity] generated in the first stage of our \augmentation{} data generation pipeline to infer the appropriate feedback timestamp.

\begin{tcolorbox}[
    colback=blue!5!white, 
    colframe=blue!75!black, 
    title=\textbf{Feedback Timestamp (Greater Counterfactual Quantity)},
    fonttitle=\bfseries,
    coltitle=black, 
    colbacktitle=blue!20!white, 
    sharp corners,
    boxrule=0.5pt
]
You are an expert cooking assistant. You are watching a person follow a step by step recipe.

\hrulefill 

\textbf{\#\#INSTRUCTIONS:}\\
You have provided the person with the following instruction: [org\_instruction].  \\

You are now provided with the following step by step account of the person's activities, along with the corresponding timestamps: [step\_by\_step\_instruction].\\

Your task is to find the timestamp at which it becomes apparent that the person uses less than the instructed quality of the following: [ingredient]. The person was instructed to use the following quantity [quantity] but instead used [counterfactual\_quantity]. \\

WAIT and make sure that the person does not intend to use more (the correct quantity). \\

Choose one of the following options: [timestamps]. Return the timestamp in seconds from the options above as a python dict of the form: \{`timestamp': ...\}. DO NOT RETURN ANYTHING OTHER THAN THE PYTHON DICT. 
\end{tcolorbox}

\subsection{Preparation error}
In addition to the step by step descriptions [step\_by\_step\_descriptions] at timestamps [timestamps], we use the attributes: [current\_prep\_method], [object] extracted from the recipe step and the counterfactual preparation method  [alternative\_prep\_method] generated in the first stage of our \augmentation{} data generation pipeline. 

\begin{tcolorbox}[
    colback=blue!5!white, 
    colframe=blue!75!black, 
    title=\textbf{Feedback Timestamp},
    fonttitle=\bfseries,
    coltitle=black, 
    colbacktitle=blue!20!white, 
    sharp corners,
    boxrule=0.5pt
]
You are an expert cooking assistant. You are watching a person follow a step by step recipe.

\hrulefill 

\textbf{\#\#INSTRUCTIONS:}\\
You have provided the person with the following instruction: [org\_instruction].  \\

You are now provided with the following step by step account of the person's activities, along with the corresponding timestamps: [step\_by\_step\_instruction].\\

Your task is to find the timestamp at which it becomes apparent that the person has used the following (wrong) preparation method: [current\_prep\_method] -- on the food item or cooking utensil: [object] -- instead of the correct preparation method: [alternative\_prep\_method]. \\

Choose one of the following options: [timestamps]. Return the timestamp in seconds from the options above as a python dict of the form: \{`timestamp': ...\}. DO NOT RETURN ANYTHING OTHER THAN THE PYTHON DICT. 
\end{tcolorbox}

\subsection{Technique error}
In addition to the step by step descriptions [step\_by\_step\_descriptions] at timestamps [timestamps], we use the attributes: [technique] and [item], extracted from the recipe step and the counterfactual technique  [alternative\_technique] generated in the first stage of our \augmentation{} data generation pipeline. 

\begin{tcolorbox}[
    colback=blue!5!white, 
    colframe=blue!75!black, 
    title=\textbf{Feedback Timestamp},
    fonttitle=\bfseries,
    coltitle=black, 
    colbacktitle=blue!20!white, 
    sharp corners,
    boxrule=0.5pt
]
You are an expert cooking assistant. You are watching a person follow a step by step recipe.

\hrulefill 

\textbf{\#\#INSTRUCTIONS:}\\
You have provided the person with the following instruction: [org\_instruction].  \\

You are now provided with the following step by step account of the person's activities, along with the corresponding timestamps: [step\_by\_step\_instruction].\\

Your task is to find the timestamp at which it becomes apparent that the person is using the wrong technique: [technique] -- and the person does not intend to use the correct technique: [alternative\_technique].  \\

Choose one of the following options: [timestamps]. Return the timestamp in seconds from the options above as a python dict of the form: \{`timestamp': ...\}. DO NOT RETURN ANYTHING OTHER THAN THE PYTHON DICT. 
\end{tcolorbox}

\subsection{Temperature error}
In addition to the step by step descriptions [step\_by\_step\_descriptions] at timestamps [timestamps], we use the attributes: [heat\_setting], [kitchen\_item], [food\_item] extracted from the recipe step and the counterfactual heat setting [counterfactual\_heat\_setting] generated in the first stage of our \augmentation{} data generation pipeline. 

\begin{tcolorbox}[
    colback=blue!5!white, 
    colframe=blue!75!black, 
    title=\textbf{Feedback Timestamp},
    fonttitle=\bfseries,
    coltitle=black, 
    colbacktitle=blue!20!white, 
    sharp corners,
    boxrule=0.5pt
]
You are an expert cooking assistant. You are watching a person follow a step by step recipe.

\hrulefill 

\textbf{\#\#INSTRUCTIONS:}\\
You have provided the person with the following instruction: [org\_instruction].  \\

You are now provided with the following step by step account of the person's activities, along with the corresponding timestamps: [step\_by\_step\_instruction].\\

Your task is to find the timestamp at which it becomes apparent that the person uses the wrong heat/temperature setting: [heat\_setting] -- of the utensil/appliance: [kitchen\_item] -- to cook the following: [food\_item]-- instead of the correct heat/temperature setting: [counterfactual\_heat\_setting].   \\

Choose the EARLIEST timestamp where it becomes clear that the person has chosen the incorrect heat/temperature setting. (This is usually when the person starts an appliance or adjusts the heat setting).\\

Choose one of the following options: [timestamps]. Return the timestamp in seconds from the options above as a python dict of the form: \{`timestamp': ...\}. DO NOT RETURN ANYTHING OTHER THAN THE PYTHON DICT. 
\end{tcolorbox}

\subsection{Timing error}
In addition to the step by step descriptions [step\_by\_step\_descriptions] at timestamps [timestamps], we use the attributes: [amount\_of\_time] and [kitchen \_item] extracted from the recipe step and the counterfactual amount of time [counterfactual\_amount\_of\_time] generated in the first stage of our \augmentation{} data generation pipeline. 
We use two different prompts for [counterfactual\_amount\_of\_time] being larger or smaller than [amount\_of\_time] in the original recipe step.

\begin{tcolorbox}[
    colback=blue!5!white, 
    colframe=blue!75!black, 
    title=\textbf{Feedback Timestamp (Smaller Counterfactual Duration)},
    fonttitle=\bfseries,
    coltitle=black, 
    colbacktitle=blue!20!white, 
    sharp corners,
    boxrule=0.5pt
]
You are an expert cooking assistant. You are watching a person follow a step by step recipe.

\hrulefill 

\textbf{\#\#INSTRUCTIONS:}\\
You have provided the person with the following instruction: [org\_instruction].  \\

You are now provided with the following step by step account of the person's activities, along with the corresponding timestamps: [step\_by\_step\_instruction].\\

First find the [start\_timestamp] when the person starts to cook: [food\_item] -- using the utensil/appliance: [kitchen\_item]. Now, instead of the specified amount of time: [amount\_of\_time] -- the person cooks for: [counterfactual\_amount\_of\_time]. Find the timestamp [critical\_timestamp] at which it becomes apparent that the person does not cook: [food\_item] -- using the utensil/appliance: [kitchen\_item] -- for long enough.   \\

Choose one of the following options: [timestamps]. Choose the LAST timestamp where it becomes clear that the person has not cooked for long enough, after the person stops cooking [critical\_timestamp] with the [kitchen\_item], at least [amount\_of\_time] from the beginning of the cooking process [start\_timestamp].\\

Return the timestamps in seconds from the options above as a python dict of the form: \{`start\_timestamp': ..., `critical\_timestamp': ...\}. The `start\_timestamp' and `critical\_timestamp' should be at least [amount\_of\_time] seconds apart. If you are not sure of the [start\_timestamp] or [critical\_timestamp] return None. DO NOT RETURN ANYTHING OTHER THAN THE PYTHON DICT.
\end{tcolorbox}

\begin{tcolorbox}[
    colback=blue!5!white, 
    colframe=blue!75!black, 
    title=\textbf{Feedback Timestamp (Larger Counterfactual Duration)},
    fonttitle=\bfseries,
    coltitle=black, 
    colbacktitle=blue!20!white, 
    sharp corners,
    boxrule=0.5pt
]
You are an expert cooking assistant. You are watching a person follow a step by step recipe.

\hrulefill 

\textbf{\#\#INSTRUCTIONS:}\\
You have provided the person with the following instruction: [org\_instruction].  \\

You are now provided with the following step by step account of the person's activities, along with the corresponding timestamps: [step\_by\_step\_instruction].\\

First find the [start\_timestamp] when the person starts to cook: [food\_item] -- using the utensil/appliance: [kitchen\_item]. Now, instead of the specified amount of time: [amount\_of\_time] -- the person cooks for: [counterfactual\_amount\_of\_time]. Find the timestamp [critical\_timestamp] at which it becomes apparent that the person cooks: [food\_item] -- using the utensil/appliance: [kitchen\_item] -- for too long.   \\

Choose one of the following options: [timestamps]. Choose the LAST timestamp where it becomes clear that the person has cooked for too long, about [amount\_of\_time] from the beginning of the cooking process [start\_timestamp] with the [kitchen\_item] (don't wait too long). \\

Return the timestamps in seconds from the options above as a python dict of the form: \{`start\_timestamp': ..., `critical\_timestamp': ...\}. The `start\_timestamp' and `critical\_timestamp' should be at least [amount\_of\_time] seconds apart. If you are not sure of the [start\_timestamp] or [critical\_timestamp] return None. DO NOT RETURN ANYTHING OTHER THAN THE PYTHON DICT.
\end{tcolorbox}



\end{document}